\documentclass{article} % For LaTeX2e
\usepackage{iclr2026_conference,times}

\renewcommand{\headrulewidth}{0pt}

\usepackage{siunitx}
\usepackage{booktabs}
\usepackage{array}
\usepackage{tabularx}
\usepackage{subcaption}
\usepackage{boldline}
\usepackage{multirow}
\usepackage{adjustbox}
\usepackage{colortbl}
\usepackage{graphicx}
\usepackage{amsmath}
\usepackage{amssymb}
\usepackage{hyperref}
\usepackage{url}
\usepackage{placeins}
\usepackage[most]{tcolorbox}
\usepackage{enumitem}
\usepackage{float}

\newcommand{\Description}[1]{}

\usepackage{titlesec}
\titlespacing*{\section}{0pt}{6pt plus 1pt minus 1pt}{3pt plus 1pt minus 1pt}
\titlespacing*{\subsection}{0pt}{4pt plus 1pt minus 1pt}{2pt plus 1pt minus 1pt}
\titlespacing*{\paragraph}{0pt}{3pt plus 1pt minus 1pt}{2pt}

\definecolor{darkfuchsia}{HTML}{C90076}
\newcommand{\qtitle}[1]{\textcolor{darkfuchsia}{#1}}

\newtcolorbox{answerbox}[1]{
  colback=darkfuchsia, colframe=darkfuchsia,
  coltext=white, boxrule=0pt, arc=2pt,
  left=5pt, right=5pt, top=2pt, bottom=2pt,
  before skip=3pt, after skip=3pt,
  fontupper=\small,
  fonttitle=\bfseries\small, title={#1}
}

\iclrfinalcopy % Uncomment only for the camera-ready version — NOT for double-blind submission.

\title{Do You Even Need More Data?\\
Rethinking LLM Deployment for Label-Scarce Tabular Classification with TabLens}

\author{Daria Grushina$^{1,2,3}$, Kseniia Kuvshinova$^{4}$, Alina Kostromina$^{1,2}$,\\
\textbf{Aziz Temirkhanov$^{1,2}$, Mile Mitrovic$^{1}$, Dmitry Simakov$^{1}$}\\[4pt]
$^{1}$Sb AI Lab, Russia \quad $^{2}$HSE University, Russia \quad $^{3}$NES, Russia\\
$^{4}$MBZUAI, United Arab Emirates\\[2pt]
\small\texttt{dgrushina@nes.ru, Kseniia.Kuvshinova@mbzuai.ac.ae,}\\
\small\texttt{alina.kostromina@gmail.com, temirkhanovmail@gmail.com,}\\
\small\texttt{milemitrovic888@gmail.com, dmitryevsimakov@gmail.com}}

\begin{document}

\maketitle

\begin{abstract}
Every tabular classification task requires sufficient labeled data. When labels are scarce, practitioners can collect more examples, encode domain expertise into features, or accept lower predictive performance. Large language models offer a potential alternative by leveraging knowledge acquired during pretraining. This possibility raises three distinct questions: (i) can zero-shot inference reduce the need for labeled data; (ii) can a richer prompt reduce the need for collecting more examples; (iii) and can encoded domain knowledge compensate for missing data? To answer each question under explicit, testable conditions, we introduce TabLens, a benchmark of 91 real-world and synthetic datasets across eight domains, together with eight industrial datasets containing free-text columns.

First, below a complexity threshold that can be estimated from a dataset's intrinsic dimensionality before running an LLM, zero-shot inference matches the accuracy that would otherwise require up to approximately 16 labeled examples to collect. This holds whether the model generates a written answer or, more cheaply, scores candidate labels directly from their log-probabilities -- the latter loses almost no accuracy except for the smallest models. Above the threshold, this advantage disappears. However, on datasets with free-text columns, zero-shot LLMs are even better than 16-shot TabPFN-3 Plus, and no longer require manual feature engineering. Second, a richer prompt reduces the number of required demonstrations only when it includes both the real column names and a plain-language task description. Providing only one of these sources of context can perform worse than providing neither, particularly at higher shot counts. Third, on the LLM-synthetic datasets, explicit if--then expert decision rules improve performance only when the prompt provides sufficient task context for the model to interpret them. Even then, they only substitute for up to about 16 real examples, after which collecting more data is more effective. Together, these results establish testable conditions under which LLMs can reduce data-collection and feature-engineering requirements for tabular classification.
\end{abstract}

% =====================================================================
\section{Introduction}
\label{sec:intro}

Tabular classification is widely used in healthcare, finance, and marketing~\cite{wang2025towards}. Although gradient boosting methods~\cite{chen2016xgboost, prokhorenkova2018catboost, ke2017lightgbm} and specialized neural architectures~\cite{chen2022danets, arik2021tabnet, huang2012tabtransformer, gorishniy2410tabm} perform well with sufficient labeled data, they remain limited in low-data regimes and cross-task generalization.

In-context learning methods such as TabPFN~\cite{hollmann2022tabpfn} and TabICL~\cite{qu2025tabicl} address data scarcity through large-scale synthetic pretraining, but still require labeled examples from the target task. Large Language Models (LLMs), by contrast, can classify from task descriptions and, optionally, a few demonstrations without parameter updates~\cite{hegselmann2023tabllm, jaitly2023towards}. However, their behavior on tabular data remains inconsistent~\cite{slack2023tablet, manikandan2023language, cheng2025tabfsbench}, and the respective roles of model priors, task descriptions, external knowledge, and demonstrations remain unclear.

Whether large language models are good at tabular classification is too coarse a question to build a benchmark around, since any answer is meaningful only relative to the alternative a practitioner would otherwise use. In practice, a team facing scarce labeled tabular data has to pay for one of a few things: collecting more labeled examples, writing richer prompts and task descriptions, or hand-coding domain expertise as rules. The question we actually care about is which of these costs an LLM lets a practitioner skip, and under what conditions. We turn this into three concrete questions:

\begin{itemize}[leftmargin=*, itemsep=2pt]
    \item[\textbf{\qtitle{Q1}}] \textbf{Can zero-shot inference substitute for collecting labeled data?}
    \item[\textbf{\qtitle{Q2}}] \textbf{Can richer prompt context substitute for collecting additional examples?}
    \item[\textbf{\qtitle{Q3}}] \textbf{Can encoded expert knowledge substitute for a lack of data?}
\end{itemize}

We introduce TabLens to answer each question. For each one we report a concrete, checkable condition that a researcher or data scientist can test against their own data, rather than a single yes-or-no verdict. The benchmark isolates the information sources involved in each substitution -- model priors, task descriptions, demonstrations, external knowledge, and inference mode -- and relates their usefulness to a TabPFN-based measure of dataset complexity. We further extend the core benchmark with a dedicated suite of 8 industrial \emph{text-tabular} datasets, letting us test whether the same substitutions hold once part of the input is free-form natural-language text rather than structured columns. As an ablation on \qtitle{Q1}, we additionally ask whether the zero-shot-for-data substitution requires costly autoregressive generation, or whether cheap forward scoring suffices.

Our results show that each substitution has a clean boundary, developed in full in Sections~\ref{sec:q1}--\ref{sec:q3}: zero-shot inference substitutes for data collection only below a predictable complexity threshold (strengthening on free-form text), and does so under either inference mode -- cheap forward scoring can replace autoregressive generation for mid-to-large open-weight models, though not for small models or pure zero-shot use; prompt context substitutes for demonstrations only when complete, and can hurt when partial; and encoded expert knowledge substitutes for data only alongside a matching prior, and only up to $\sim$16 shots.

Our main contributions are:
\begin{itemize}
\item We introduce TabLens, a benchmark for zero-shot and few-shot in-context tabular classification with LLMs, comprising 91 core datasets across eight domains plus an 8-dataset text-tabular extension.
\item We answer three substitution questions -- data collection, additional demonstrations, and expert knowledge -- each with a controlled ablation, statistical significance testing, and an explicit boundary condition rather than an aggregate score.
\item We show that the zero-shot-for-data substitution is governed by a measurable, TabPFN- and intrinsic-dimensionality-predictable complexity threshold, that it extends to feature collection itself on datasets carrying free-form text, and that it holds under cheap forward scoring as well as full autoregressive generation for mid-to-large open-weight models.
\item We show that context and demonstrations are partial substitutes that can interfere destructively when only one is partially supplied, and that expert knowledge requires a contextual prior to be useful at all.
\end{itemize}

\section{Related Work}
\label{sec:related}

Existing tabular benchmarks such as OpenML-CC18~\cite{bischl2021openml}, TabZilla~\cite{mcelfresh2023neural}, and TabArena~\cite{erickson2025tabarena} improve evaluation consistency but primarily target supervised learning. They are not designed to isolate in-context learning factors such as prompt format, task descriptions, context budgets, or demonstration selection.

Recent studies evaluate LLMs on tabular prediction more directly. TABLET~\cite{slack2023tablet} reports benefits from natural-language instructions alongside limitations in instruction faithfulness, while TabuLa-8B~\cite{gardner2024large} reports zero- and few-shot transfer after large-scale tabular language-model training; TabFSBench~\cite{cheng2025tabfsbench} independently documents that LLM tabular performance is brittle under feature shifts. However, \cite{gorla2026illusion} argues that apparent generalization may be inflated by task aggregation, leakage, contamination, or format familiarity. Similarly, \cite{silvestri2025evaluating} finds that LLMs may retain latent knowledge of public datasets when semantic column names and interpretable values are preserved, and \cite{bordt2024elephants} documents memorization and learning of tabular data in LLMs more broadly. These results motivate protocols that separate genuine in-context generalization from contamination and semantic leakage, which is why we complement our aggregate comparisons (Section~\ref{sec:q1}) with explicit recognition and reconstruction probes.

Two dedicated LLM-tabular pipelines are directly relevant to our design: FeatLLM~\cite{han2024featllm} distills LLM-generated features for a downstream few-shot classifier, and Knauer et al.~\cite{knauer2025llmtrees} induce a zero-shot decision tree from a task description alone. We use both as baselines in Section~\ref{sec:q1_h1a} rather than only citing them, since they let us test directly whether a dedicated pipeline beats plain prompting.

Benchmarks such as Table Meets LLM~\cite{sui2023gpt4table} and TableBench~\cite{wu2025tablebench} instead focus on table understanding and question answering rather than row-wise classification. A separate line of work targets \emph{joint} text-tabular inputs, where one or more columns carry free-form natural language alongside structured features, including multimodal AutoML for tabular data with text fields~\cite{shi2021benchmarking}, curated multimodal benchmarks exploiting text--tabular synergy~\cite{arazi2026multabench}, and text meta-feature analyses of tabular learning~\cite{blayer2026strable, mraz2025towards}. These resources focus on dataset construction and tabular foundation models rather than on when zero-shot LLMs specifically become competitive; TabLens's text-tabular extension (Section~\ref{sec:q1_text}) targets this gap directly by decomposing text-tabular datasets into text-only and tabular-only variants and tracking the zero-shot LLM vs.\ label-aware baseline comparison across that decomposition.

\section{TabLens Overview}
\label{sec:3}

\paragraph{Terminology}
Table~\ref{tab:glossary} defines the recurring prompt and evaluation components used without redefinition from Section~\ref{sec:q1} onward.

\begin{table}[H]
\centering
\caption{Key terms used throughout the paper.}
\label{tab:glossary}
\tiny
\renewcommand{\arraystretch}{1.15}
\setlength{\tabcolsep}{3pt}
\begin{tabularx}{\linewidth}{@{}
    >{\raggedright\arraybackslash\hsize=.62\hsize}X
    >{\raggedright\arraybackslash\hsize=1.38\hsize}X
    >{\raggedright\arraybackslash\hsize=.62\hsize}X
    >{\raggedright\arraybackslash\hsize=1.38\hsize}X @{}}
\hline
\textbf{Term} & \textbf{Meaning} & \textbf{Term} & \textbf{Meaning} \\
\hline
Zero-shot & Classification with $k=0$ labeled prompt demonstrations. &
Full / complete context & Schema indicators and task context are both present. \\
Few-shot ($k$-shot) & $k$ labeled rows precede the query. &
Partial / incomplete context & Only schema indicators or task context are present. \\
Prior knowledge / prior & Knowledge encoded during pretraining, before task-specific text. &
Expert knowledge (rules) & Explicit if--then rules supplied as $r$ (Sec.~\ref{sec:q3}). \\
Task context (prior context) & Description of the domain, objective, and label meaning, denoted $c$ (Sec.~\ref{sec:q3}). &
Forward scoring & Target-label log-probabilities read in one forward pass (Sec.~\ref{sec:q1_h1d}). \\
Schema indicators & Human-readable column names rather than generic placeholders. &
Generation & Autoregressive label generation followed by parsing. \\
Intrinsic dimensionality & Model-free feature-matrix complexity score (Sec.~\ref{sec:q1_h1b}). &
Substitution & Replacing a deployment cost---data, examples, or expertise---with an LLM capability under a testable condition. \\
\hline
\end{tabularx}
\end{table}

TabLens evaluates low-data binary tabular classification while isolating serialization, example count, task knowledge, dataset complexity, and input modality.

\paragraph{Dataset suites}
TabLens contains \textbf{99} datasets across five suites (Table~\ref{tab:llmtabbench_suites}). The core suite comprises \textbf{24} public OpenML~\cite{bischl2021openml} and Kaggle datasets from \textbf{8} domains. All datasets use a fixed class-balanced stratified split with a $0.3$ test fraction,\footnote{This reproduces the train/test partition of the $k$-fold cross-validation split used by TabArena~\cite{erickson2025tabarena}.} while each text-tabular dataset is additionally decomposed into \emph{full}, \emph{text-only}, and \emph{tabular-only} variants to isolate free-form text from structured features.

\begin{table}[t]
\centering
\caption{TabLens dataset suites and associated hypotheses.}
\label{tab:llmtabbench_suites}
\tiny
\renewcommand{\arraystretch}{1.15}
\setlength{\tabcolsep}{3pt}
\begin{tabularx}{\linewidth}{@{}
    >{\raggedright\arraybackslash\hsize=.4\hsize}X
    r
    >{\raggedright\arraybackslash\hsize=1.3\hsize}X
    >{\raggedright\arraybackslash\hsize=.3\hsize}X @{}}
\hline
\textbf{Suite} & \textbf{\#} & \textbf{Purpose / construction} & \textbf{Used by} \\
\hline
Core real-world & 24 &
Public OpenML/Kaggle datasets grouped into 8 domains. & \qtitle{H1a}, \qtitle{H2a}--d \\
Temporal (post-cutoff) & 8 &
Datasets released after Oct.~2024 to reduce pretraining-contamination risk and probe leakage. & \qtitle{H1a}, \qtitle{H1d} \\
LLM-synthetic & 24 &
LLM-generated counterparts of the real datasets, each with explicit decision rules. & \qtitle{H3} \\
Complexity grid (MLP-prior) & 35 &
Controlled difficulty: 7 levels $\times$ 5 random seeds. & \qtitle{H1b} \\
Text-tabular (industrial) & 8 &
Mixed free-text and structured columns, decomposed into full/text-only/tabular-only variants. & \qtitle{H1c}, \qtitle{H2d} \\
\hline
\textbf{Total} & \textbf{99} & & \\
\hline
\end{tabularx}
\end{table}

\paragraph{Evaluation axes}
We evaluate zero-shot and 4, 8, 16, 32, and 64-shot inference using five serializations: feat\_val, feat\_val\_mask, markdown, markdown\_mask, and HTML. The complexity grid uses the two masked formats described in the appendix. Context is provided throughout; the synthetic suite additionally includes context-free and knowledge-augmented prompts with LLM-generated decision rules. Every few-shot setting uses five independent stratified demonstration draws.

\paragraph{Models and baselines}
We evaluate GPT-4o-mini, Qwen3-1.7B, Qwen3-8B, Qwen3-14B~\cite{yang2025qwen3}, and GLM-4.7-Flash; GLM has partial suite coverage, documented with its interpretation constraints in Appendix~\ref{app:glm_coverage}. For GLM and Qwen, we compare normalized target-label \emph{forward scoring} with parsed free-form \emph{generation}. Baselines are TabPFN-3 Plus~\cite{grinsztajn2026tabpfn3}, TabPFN-2.5 Plus~\cite{priorlabs2025tabpfn3}, TabICL~\cite{qu2025tabicl}, XGBoost~\cite{chen2016xgboost}, CatBoost~\cite{prokhorenkova2018catboost}---including native text features on the text-tabular suite---Random Forest, Logistic Regression, $k$-nearest neighbors, and majority-class prediction, all evaluated under the same protocol. On the 32 real datasets, we also evaluate FeatLLM~\cite{han2024featllm} and zero-shot LLM decision-tree induction~\cite{knauer2025llmtrees}; such methods primarily target few-shot learning, while pure zero-shot alternatives remain uncommon.

\paragraph{Scoring}
We report ROC--AUC from continuous scores because it measures positive-over-negative ranking independently of deployment-specific thresholds and error costs. Following standard ROC practice, values below $0.5$ indicate a ranking that is systematically inverted rather than uninformative, since the classifier's decisions could simply be negated to obtain an equally strong classifier with AUC $1-\mathrm{AUC}$~\citep{fawcett2006introduction}; we therefore replace any such score by $1-\mathrm{AUC}$. Results are averaged over five demonstration draws.

\paragraph{Testing zero-shot performance against random guessing}
For H1a, the dataset---not an individual prediction or seed---is the experimental unit. For each model and each of the 24 classic datasets, we average zero-shot ROC--AUC over up to five available runs in the primary full-context generation configuration and compute $d_j=\overline{\mathrm{AUC}}_j-0.5$. We test $H_0:\operatorname{median}(d_j)\leq0$ against $H_1:\operatorname{median}(d_j)>0$ using a one-sided Wilcoxon signed-rank test, the standard non-parametric recommendation for across-dataset comparisons~\cite{demsar2006statistical}. Holm's step-down procedure controls the family-wise error rate across the five model-level tests~\cite{holm1979simple}; we report the median paired difference and matched-pairs rank-biserial correlation as effect sizes. This protocol requires only dataset-level summary ROC--AUC values, not stored per-example predictions.

\section{\qtitle{Q1: Can Zero-Shot LLM Inference Replace Collecting Labeled Data?}}
\label{sec:q1}

We first ask whether zero-shot LLM inference can match few-shot baselines without using labeled examples from the target task. We test this claim along three dimensions: overall competitiveness against a strong few-shot baseline (H1a), the existence of a predictable complexity boundary (H1b), and robustness across input modalities, including text-tabular data (H1c). Finally, we run a fourth ablation: whether the results, and respective computational costs, depend on the inference mode (autoregressive generation and direct label scoring) (H1d).

\subsection{\qtitle{H1a: Zero-Shot LLMs vs.\ $k$-Shot TabPFN-3 Plus}}
\label{sec:q1_h1a}

\textbf{Setup} \label{sec:4rq1}
We evaluate zero-shot classification with 5 LLMs on the 24 core real-world
datasets plus 8 post-cutoff datasets (32 total). On the 24 classic datasets,
we test whether each model exceeds random guessing ($\mathrm{ROC\mbox{-}AUC}=0.5$)
with a one-sided Wilcoxon signed-rank test over dataset-level mean ROC--AUC
differences, followed by Holm correction across the five models
~\cite{demsar2006statistical,holm1979simple}; this requires no per-example
predictions (full protocol and visualization in Appendix~\ref{app:chance_test}). We split the pool into
\emph{Classic} (the 24 established benchmarks) and \emph{New} to
probe memorization. \emph{New} suite contains 8 datasets
published after October 2024, unavailable to any model's pretraining. The sole exception is GLM 4.7 Flash: as a recently released model, it is impossible at present to assemble a sufficient number of new datasets satisfying our criteria that would also postdate its pretraining. 

\begin{table*}[!htbp]
\sisetup{separate-uncertainty=true,table-align-uncertainty=true,table-figures-uncertainty=1}
\centering
\scriptsize
\setlength{\tabcolsep}{2pt}
\resizebox{\textwidth}{!}{%
\begin{tabular}{lcccccc}
\toprule
Dataset Group & GPT-4o-mini (gen) & Qwen3-1.7B (gen) & Qwen3-8B (gen) & Qwen3-14B (gen) & GLM-4.7 Flash (gen) & TabPFN-3 Plus (16-shot) \\
\midrule
All & \num{0.722 \pm 0.008} & \num{0.605 \pm 0.007} & \num{0.684 \pm 0.001} & \num{0.693 \pm 0.000} & \num{0.649 \pm 0.028} & \num{0.732 \pm 0.062} \\
Classic & \num{0.731 \pm 0.009} & \num{0.608 \pm 0.007} & \num{0.689 \pm 0.000} & \num{0.699 \pm 0.000} & \num{0.667 \pm 0.030} & \num{0.736 \pm 0.060} \\
New & \num{0.664 \pm 0.008} & \num{0.587 \pm 0.005} & \num{0.657 \pm 0.001} & \num{0.657 \pm 0.000} & \num{0.607 \pm 0.023} & \num{0.693 \pm 0.073} \\
\bottomrule
\end{tabular}
}%
\caption{Zero-shot LLMs vs.\ TabPFN-3 Plus (16-shot), broken down by dataset group: \emph{All} (32 datasets), \emph{Classic} (24 established benchmarks), and \emph{New} (8 post-cutoff datasets).}
\label{tab:h1a_zero_shot_vs_tabpfn_all}
\end{table*}

The corresponding per-domain comparison is reported in
Figure~\ref{fig:rq1_radar} (Appendix~\ref{app:key_figures}).

\textbf{Results}
On the full 32-dataset pool (Table~\ref{tab:h1a_zero_shot_vs_tabpfn_all}), every mid-to-large
model approaches the performance of 16-shot TabPFN-3 Plus ($0.753$): GPT-4o-mini
$0.712$, Qwen3-8B $0.698$, Qwen3-14B $0.714$, GLM-4.7 Flash $0.649$. Only the
smallest model falls well short (Qwen3-1.7B, $0.626$). The dataset-level
differences from random guessing and the Holm-adjusted Wilcoxon results are
shown in Figure~\ref{fig:q1_zero_shot_vs_random} (Appendix~\ref{app:rq1}). Splitting \emph{Classic}
vs.\ \emph{New} moves every model, TabPFN-3 Plus included, down by a similar
$0.05$--$0.08$ (e.g.\ TabPFN-3 Plus $0.772 \to 0.695$) -- a broadly shared drop
consistent with the \emph{New} datasets simply being harder, not with LLMs
selectively losing a memorized advantage. Two robustness checks reinforce
this reading without changing it: a two-step memorization probe (recognition
+ cell reconstruction; Appendix~\ref{app:rq1}) flags only 2 of 24 classic
datasets as \textsc{likely-leaked}, and\label{sec:q1_pipelines} direct prompting matches or exceeds
two dedicated LLM-tabular pipelines, FeatLLM~\cite{han2024featllm} and LLM
Trees~\cite{knauer2025llmtrees}, at every shot count
(Table~\ref{tab:gpt_methods_vs_shots}, Appendix~\ref{app:rq1}).

\textbf{Takeaway}
For mid-to-large models, zero-shot performance is comparable to that of TabPFN-3 Plus using 16 labeled examples; only the smallest model needs the data. The performance gap observed on post-cutoff datasets is more consistent with differences in dataset difficulty than with memorization.

\subsection{\qtitle{H1b: A Predictable Complexity Threshold}}
\label{sec:q1_h1b}

If H1a holds only below a certain level of dataset complexity, practitioners need an LLM-independent criterion for determining whether a given dataset falls within this regime.

\textbf{Setup} We study controlled synthetic datasets generated from an
MLP-based TabPFN-style prior~\cite{hollmann2022tabpfn}, with complexity
varied through network depth, width, and hidden state size across seven
difficulty levels and five random seeds (35 datasets total). For each level
we compute an intrinsic-dimensionality score, averaged over several
estimators from \texttt{scikit-dimension}~\cite{bac2021scikit,
fukunaga1971algorithm, mirkes2020fractional, fan2010intrinsic,
levina2004maximum, facco2017estimating, carter2009local, gardner2406large}
(full protocol in Appendix~\ref{app:protocol}).

\begin{figure*}[t]
    \centering
    \includegraphics[width=0.95\textwidth]{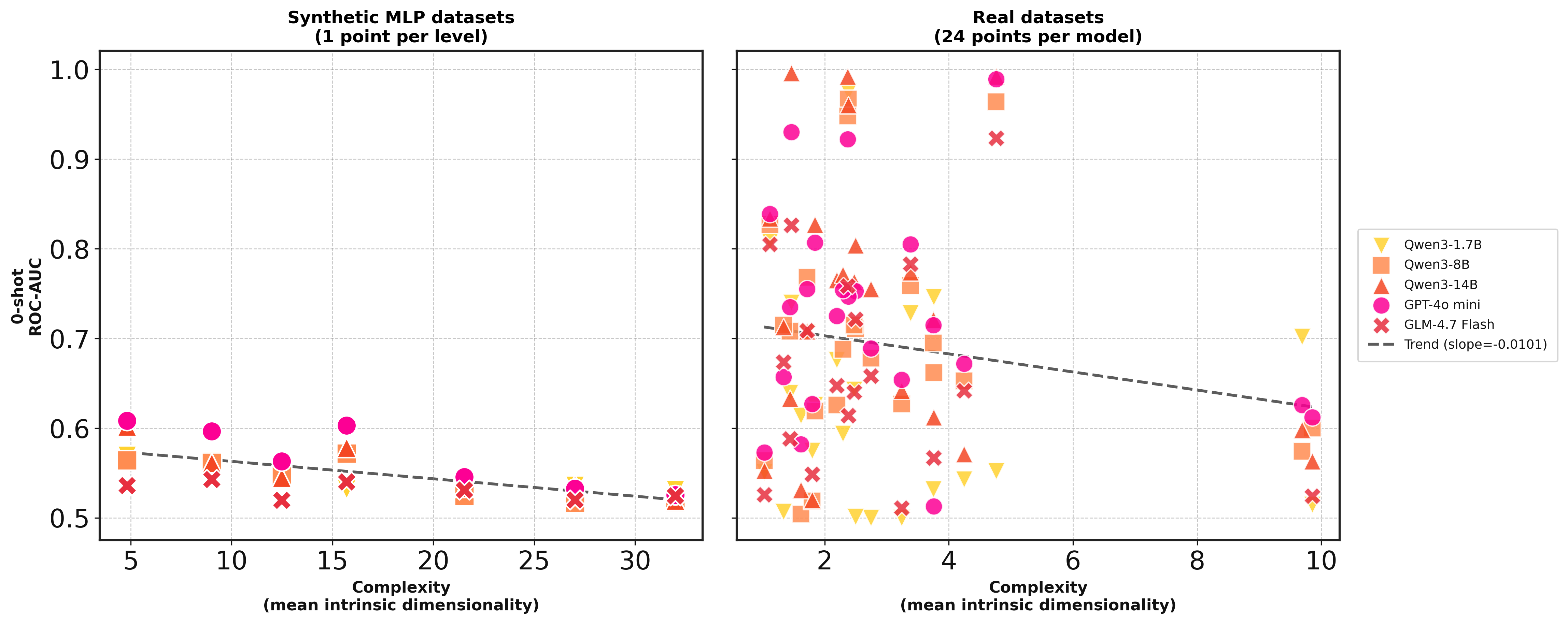}
    \caption{Intrinsic dimensionality vs.\ zero-shot ROC--AUC on the
    synthetic MLP suite and the real-world datasets.}
    \label{fig:h1b_complexity}
    \Description{Zero-shot model performance as a function of intrinsic dataset complexity.}
\end{figure*}

\textbf{Results}
Intrinsic dimensionality rises steadily from $4.8$ (MLP0) to $32.0$ (MLP9;
Table~\ref{tab:h1b_mlp_complexity}, Appendix~\ref{app:rq6}), and zero-shot
ROC-AUC falls almost in lock-step: Qwen3-14B drops from $0.602$ to
$0.516$ and GPT-4o-mini from $0.608$ to $0.526$, with nearly the entire
decline already realized by MLP1 (dimensionality $9.0$: Qwen3-14B $0.561$,
GPT-4o-mini $0.596$) and only noise-level movement from MLP2 onward. The
same ordering appears independently in 16-shot TabPFN-3 Plus's own
performance on this grid (Figure~\ref{fig:rq6_tabpfn_mlp}, Appendix), and
DeLong's test confirms a significant few-shot gain only at MLP0
(Appendix~\ref{app:rq6}).

\textbf{Takeaway}
A dataset's intrinsic dimensionality -- computable without ever calling an
LLM -- locates it on the same complexity scale where zero-shot performance
collapses. In practice: below roughly dimensionality $9$, zero-shot inference is most promising; above it, collect the data instead.

\subsection{\qtitle{H1c: The Text-Tabular Niche}}
\label{sec:q1_text}

H1a and H1b are both established on structured tabular columns. Many
industrial datasets, however, mix structured columns with free-form
natural-language fields (product reviews, job descriptions, clinical notes).
We ask whether the zero-shot-for-data substitution not only holds but
\emph{strengthens} once part of the signal is carried by text that
label-aware tabular learners cannot read natively.

\textbf{Setup}
We curate 8 industrial text-tabular datasets in which one or more columns
carry free-form natural language alongside numerical and categorical
features (full descriptions in Appendix~\ref{app:texttab}), decomposing
each into a \emph{text-only} variant (structured columns removed) and a
\emph{tabular-only} variant (text columns removed). TabPFN-3 Plus and
CatBoost have no zero-shot mode, so their 0-shot cells are fixed at the
random baseline ($0.5$). The corresponding shot trajectories are reported in
Figure~\ref{fig:texttab_niche} (Appendix~\ref{app:key_figures}).

\textbf{Results}
At 0-shot (Table~\ref{tab:h1c_text_tabular_views}, Appendix~\ref{app:texttab}), GPT-4o-mini already scores
\emph{higher} on \emph{text-only} ($0.687$) than on \emph{full}
($0.677$) or \emph{tabular-only} ($0.640$) -- removing structured columns slightly improves performance. At 16 shots, TabPFN-3
Plus and CatBoost achieve above-chance text-only performance, but it is barely
above chance ($0.539$ and $0.543$): both stay $0.14$+ ROC-AUC below
GPT-4o-mini's zero-shot text-only number, and CatBoost's own \emph{full}
variant ($0.578$) still trails GPT-4o-mini's zero-shot \emph{full} score
($0.677$). The ordering (\emph{full} $>$ \emph{text-only} $>$
\emph{tabular-only} for TabPFN-3 Plus; \emph{text-only} $\geq$ \emph{full}
$>$ \emph{tabular-only} for the LLM) is set at zero shots and persists as
shots increase (Figure~\ref{fig:texttab_niche}).

\textbf{Takeaway}
Zero-shot LLMs are most competitive when the predictive signal is concentrated in free-form text, which many conventional tabular learners cannot process directly. In this setting, LLMs may reduce not only the need for labeled examples but also the need for task-specific feature engineering.

\subsection{\qtitle{H1d: Does Inference Mode Change the Picture?}}
\label{sec:q1_h1d}

H1a--H1c all use a fixed inference mode. We test whether the conclusions of H1a--H1c are robust to replacing autoregressive generation with forward scoring.

\emph{Generation} autoregressively produces a label phrase that must be
parsed, requiring multiple decoding steps and handling malformed or
off-label outputs. \emph{Forward scoring} instead obtains the two
target-label log-probabilities in a single forward pass and normalizes
them with a sigmoid, directly returning a normalized class probability without
decoding or parsing.

\textbf{Setup} We compare both modes for models that expose token-level
log-probabilities, at $k=0$ and $k=64$, across the 32 real datasets. GPT-4o-mini
is omitted because its API does not expose the required target-label
log-probabilities.

\begin{table*}[!t]
\centering
\footnotesize
\begingroup
\sisetup{
  separate-uncertainty = true,
  table-align-uncertainty = true,
  table-figures-uncertainty = 1,
}
\setlength{\tabcolsep}{2pt}
\begin{tabular}{l cc c cc c}
\toprule
& \multicolumn{2}{c}{\textbf{0-shot}} && \multicolumn{2}{c}{\textbf{64-shot}} & \\
\cmidrule(lr){2-3}
\cmidrule(lr){5-6}
\textbf{Model} & \textbf{Generation} & \textbf{Forward} & $\Delta$ & \textbf{Generation} & \textbf{Forward} & $\Delta$ \\
\midrule
Qwen3-1.7B & \num{0.6227 \pm 0.1094} & \num{0.5319 \pm 0.0644} & $-0.0908$ & \num{0.6957 \pm 0.1129} & \num{0.6374 \pm 0.1166} & $-0.0583$ \\
Qwen3-8B & \num{0.6975 \pm 0.1429} & \num{0.6530 \pm 0.1341} & $-0.0445$ & \num{0.7766 \pm 0.1299} & \num{0.7783 \pm 0.1291} & $+0.0017$ \\
Qwen3-14B & \num{0.7143 \pm 0.1549} & \num{0.7014 \pm 0.1425} & $-0.0129$ & \num{0.7886 \pm 0.1333} & \num{0.7762 \pm 0.1312} & $-0.0124$ \\
GLM-4.7 Flash & \num{0.6676 \pm 0.1284} & \num{0.6549 \pm 0.1165} & $-0.0127$ & \num{0.7507 \pm 0.1257} & \num{0.7458 \pm 0.1707} & $-0.0050$ \\
\bottomrule
\end{tabular}
\caption{Forward scoring vs.\ generation: mean ROC-AUC $\pm$ standard deviation across the 24 classic + 8 new datasets (standard full-context prompt). $\Delta = \text{forward}-\text{generation}$.}
\label{tab:h1d_forward_vs_gen}
\endgroup
\end{table*}

\textbf{Results}
At 0-shot, forward scoring trails generation by $-0.09$ ROC-AUC for
Qwen3-1.7B, shrinking to $-0.04$ for Qwen3-8B and to about $-0.01$ for
Qwen3-14B and GLM-4.7 Flash -- the gap closes steadily with model scale. At
64 shots it closes further still: $-0.06$ for Qwen3-1.7B, essentially zero
or positive for the three larger models ($+0.002$, $-0.012$, $-0.005$).
Only the smallest model keeps a non-negligible gap, and only at low shot
counts. On the efficiency side, forward scoring is
${\approx}4.5\times$ faster at 0-shot and ${\approx}1.4\times$ faster at
64-shot, since the decoding stage it eliminates matters less once the
shared few-shot prompt dominates prefill cost.

\textbf{Takeaway}
Inference mode does not change H1a--H1c's conclusions. For mid-to-large
open-weight models, forward scoring is a safe, substantially cheaper
default; only the smallest model or the pure zero-shot regime still
favors paying for generation.

\section{\qtitle{Q2: Can Adding Context to the Prompt Replace Collecting Additional Few-Shot Examples?}}
\label{sec:q2}

If a task description or richer serialization can substitute for demonstrations, a practitioner who cannot collect more labeled examples might still improve performance simply by writing a better prompt. We evaluate this hypothesis in four steps: whether the improvement attributable to a task description alone (\qtitle{H2a}), whether real demonstrations themselves have diminishing returns (\qtitle{H2b}), whether \emph{partial} context degrades performance once demonstrations are available (\qtitle{H2c}), and whether serialization format has an effect comparable to that of adding demonstrations (\qtitle{H2d}).

\subsection{\qtitle{H2a: A Task Description Closes Part of the Gap}}
\label{sec:q2_h2a}

\textbf{Setup} We vary two channels through which in-context information
reaches the model at 0-shot: \emph{Schema indicators} (real column names
vs.\ generic placeholders) and \emph{Task context} (a natural-language task
description vs.\ a neutral, uninformative instruction). Crossing the two
gives four configurations, from \emph{NoCols-NoContext} (minimal) to
\emph{Cols-Context} (maximal); full descriptions in
Appendix~\ref{app:protocol}.

\begin{figure*}[t]
    \centering
    \includegraphics[width=0.98\textwidth]{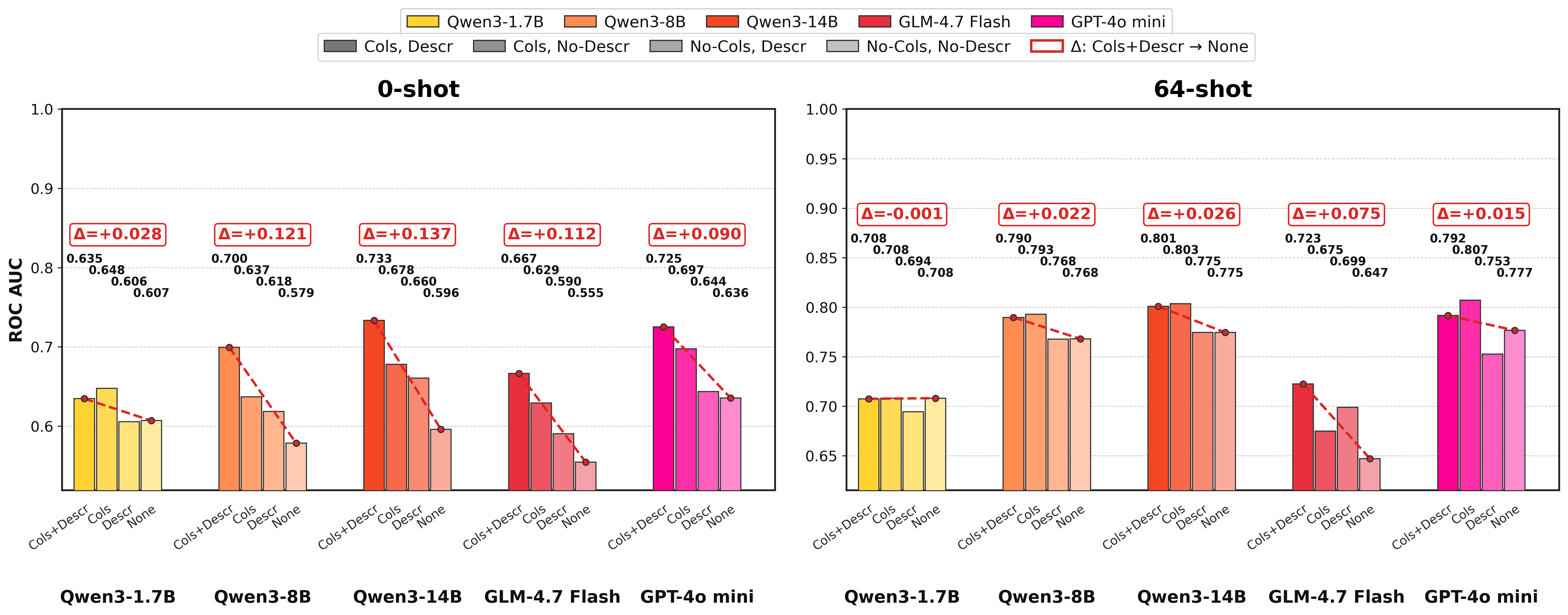}
    \caption{Context-configuration comparison on the 24 classic datasets at
    0-shot (left) and 64-shot (right). Each group shows the four
    combinations of column names and natural-language task context for one
    model; the red dashed line marks the gap between the richest
    (Cols-Context) and poorest (NoCols-NoContext) configuration.}
    \label{fig:rq2_pk0shot}
    \Description{Zero-shot and 64-shot performance across the four context configurations for five LLMs, shown side by side.}
\end{figure*}

\textbf{Results}
Moving from \emph{NoCols-NoContext} to \emph{Cols-NoContext}
(Table~\ref{tab:h2b_context_configs}, Appendix~\ref{app:rq2}) already
recovers most of the total gain for four of five models: GPT-4o-mini
$0.625\to0.682$, Qwen3-8B $0.579\to0.637$, Qwen3-14B $0.596\to0.678$,
GLM-4.7 Flash $0.564\to0.637$. Stacking the task description on top
(\emph{Cols-Context}) adds a further $0.033$--$0.055$ for those same four
models, but for Qwen3-1.7B the identical step actually costs accuracy
($0.653\to0.640$). The joint gain (\emph{NoCols-NoContext}
$\to$ \emph{Cols-Context}) is $0.10$--$0.14$ ROC-AUC for the four
larger models -- larger than either channel alone, so the two are
complementary rather than redundant.

\textbf{Takeaway} Schema indicators alone deliver most of the zero-shot
gain from in-context information; the natural-language task description
adds a further boost only at the $8$\,B parameter scale and above. This is
a \emph{partial} substitution: it recovers a meaningful fraction of what
demonstrations would provide, but, as the next two sections show, it does
not fully replace them.

\subsection{\qtitle{H2b: Real Demonstrations Have Diminishing Returns}}
\label{sec:q2_h2b}
\textbf{Setup} For $k \in \{0,4,8,16,32,64\}$ we prefix $k$ randomly sampled
labeled examples to the query and score ROC--AUC over all test instances.
Significance of each $k$-shot gain over zero-shot is assessed with DeLong's
test aggregated across the suite by Stouffer's method (Appendix~\ref{app:del_test}).

\begin{figure*}[t]
    \centering
    \includegraphics[width=0.75\textwidth]{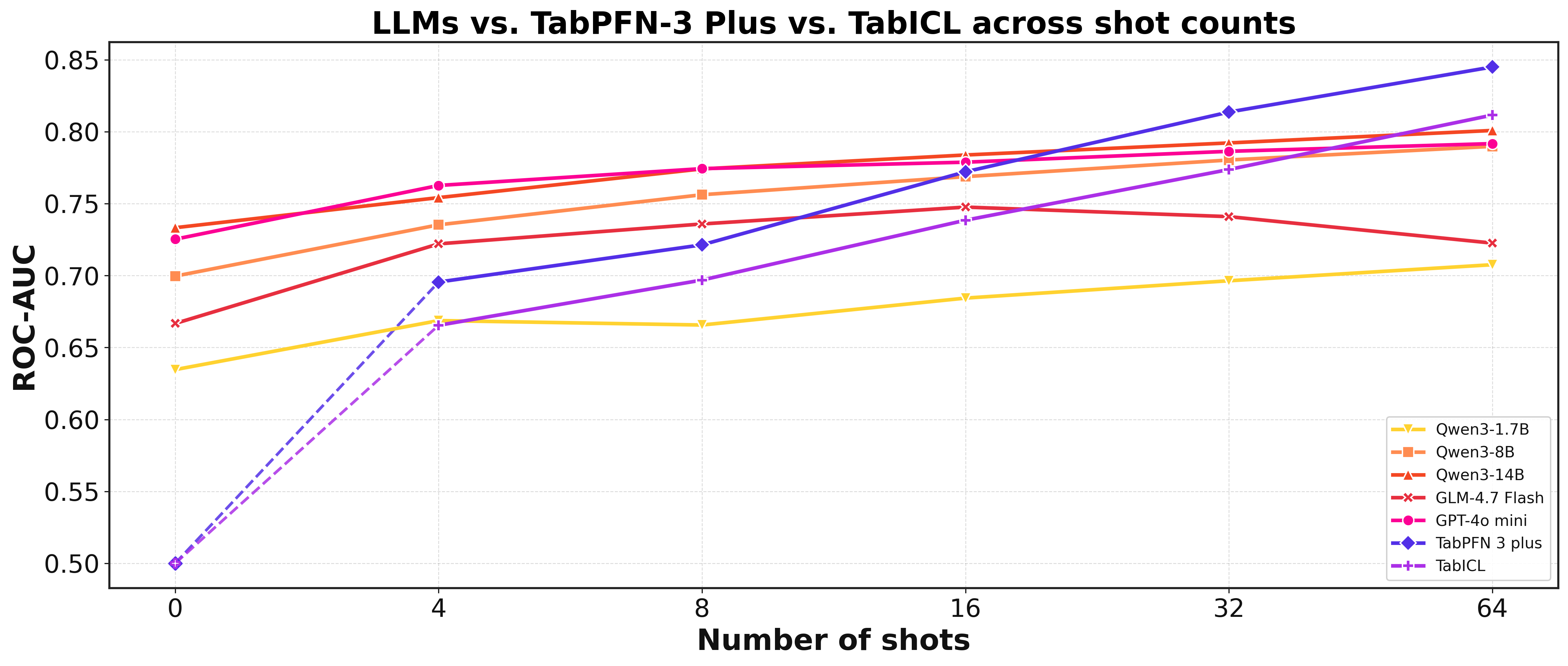}
    \caption{ROC--AUC vs.\ the number of in-context demonstrations on
    the 24 classic datasets, for all five LLMs alongside TabPFN-3 Plus and
    TabICL. TabPFN-3 Plus has no true zero-shot mode; its curve is extended
    left to chance level (0.5) as a dashed segment.}
    \label{fig:rq3_shots_curve_full}
    \Description{Performance of the evaluated LLMs, TabPFN-3 Plus, and TabICL across shot counts on the classic datasets.}
\end{figure*}

\textbf{Results}
Every LLM's per-shot gain shrinks fast (Table~\ref{tab:h2a_shot_scaling_classic}, Appendix~\ref{app:rq3}): for
Qwen3-8B the $0\to4$-shot step adds $+0.035$ ROC-AUC, but the entire
$16\to64$-shot range adds only $+0.021$ combined; for GPT-4o-mini the
$0\to4$ step alone ($+0.038$) already exceeds the $8\to64$ range combined
($+0.018$). GLM-4.7 Flash peaks at 16 shots
($0.748$) before \emph{declining} to $0.723$ by 64 shots. TabPFN-3 Plus, on
the same table, shows the opposite shape: it keeps climbing without
flattening across the same range ($0.696\to0.721\to0.772\to0.814\to0.845$
from 4 to 64 shots), overtaking every LLM's shot-saturated ceiling.

\textbf{Takeaway} LLMs do exploit demonstrations, but the marginal benefit
per shot collapses quickly and is largely exhausted by 16 shots -- unlike a
model such as TabPFN-3 Plus, which keeps improving over the same range.
Saturation here is a property of how LLMs use in-context examples, not of
the underlying task. This 16-shot budget recurs in Section~\ref{sec:q3} as
the point past which encoded expert knowledge stops adding value.

\subsection{\qtitle{H2c: Incomplete Context Can Be Worse Than No Context}}
\label{sec:q2_h2c}

\textbf{Setup} We extend the \qtitle{H2a} design with few-shot examples, drawing $k$
labeled examples per (Model, Configuration, $k$) combination and tracking
how the \emph{Cols-Context} $-$ \emph{NoCols-NoContext} gap moves as $k$
grows (Appendix~\ref{app:protocol}).

\textbf{Results}
The full-vs-minimal-context gap shrinks monotonically with $k$ for every
model that uses context at all (Table~\ref{tab:delta_context_vs_nocontext},
Appendix~\ref{app:rq4}): GPT-4o-mini falls from $+0.103$ at $k=0$ to
$+0.032$ at $k=64$; Qwen3-8B from $+0.132$ to $+0.023$; Qwen3-14B from
$+0.137$ to $+0.026$; GLM-4.7 Flash from $+0.081$ to $+0.007$. More
strikingly, once enough demonstrations are present, the \emph{complete
absence} of context (\emph{NoCols-NoContext}) can outperform the
\emph{partial}-cue configurations (\emph{NoCols-Context},
\emph{Cols-NoContext}) -- clearest for GPT-4o-mini in the 64-shot panel of
Figure~\ref{fig:rq2_pk0shot}, where the no-context, no-schema setting
overtakes its intermediate counterparts.

\textbf{Takeaway} Context and demonstrations are partial substitutes rather than additive sources of information: context mainly improves performance when demonstrations
are scarce, whereas partial context may interfere with the signal provided by demonstrations at higher shot counts.

\subsection{\qtitle{H2d: Serialization Format vs.\ Shot Count}}
\label{sec:q2_h2d}

\textbf{Setup} We rank the five serialization formats introduced with the
text-tabular suite (Section~\ref{sec:q1_text}) by zero-shot ROC-AUC per
(dataset, model) pair on the pooled 32-dataset set (Appendix~\ref{app:texttab}).

\textbf{Results} The best format (\texttt{markdown} or \texttt{html},
depending on the model) beats the worst by $1.0$--$1.5$ mean rank
positions, and masking column names costs a comparable amount regardless of
which format carries it (Table~\ref{tab:serialization_texttab},
Appendix~\ref{app:texttab}). In absolute ROC-AUC terms this is consistently
smaller than the $0.10$--$0.14$ swing from \qtitle{H2a} or the shot-driven gains
from \qtitle{H2b}, and the ordering is already set at zero shots.

\textbf{Takeaway} Serialization format is therefore a secondary optimization factor: although it can be tuned at relatively low cost, its effect is smaller than that of complete prompt context or additional demonstrations.
\section{\qtitle{Q3: Can Encoded Expert Knowledge Replace a Lack of Data?}}
\label{sec:q3}

A different kind of substitute for raw examples is explicit domain knowledge: if-then decision rules, feature-selection heuristics, or other codified expertise. We test whether supplying such rules substitutes for missing demonstrations, and under what conditions. We formalize what we mean by \emph{prior context} and \emph{expert knowledge} using the implicit Bayesian view of in-context learning~\citep{xie2022explanation}.

\subsection{\qtitle{H3: Expert Rules Need a Matching Prior}}
\label{sec:q3_h3}

\textbf{Setup} We use the LLM-synthetic suite (Table~\ref{tab:llmtabbench_suites}):
24 datasets each generated by an LLM to mirror a real-world counterpart,
together with the decision rules the same LLM produced during generation.
We evaluate all seven combinations of three in-context learning entities --
\emph{few-shot} demonstrations (16), the natural-language task description
(\emph{prior}), and the LLM-distilled decision rules (\emph{expert}; see
Appendix~\ref{app:prompt3}) -- against a 16-shot TabPFN-3 Plus baseline
($\approx 0.72$; full protocol in Appendix~\ref{app:protocol}).

\begin{table*}[t]
\sisetup{separate-uncertainty=true,table-align-uncertainty=true,table-figures-uncertainty=1}
\centering
\footnotesize
\setlength{\tabcolsep}{4pt}
\resizebox{\textwidth}{!}{%
\begin{tabular}{lccccc}
\toprule
\textbf{Prompting Regime} & \textbf{Qwen3-1.7B} & \textbf{Qwen3-8B} & \textbf{Qwen3-14B} & \textbf{GPT-4o-mini} & \textbf{GLM-4.7 Flash} \\
\midrule
Few-shot (16) & \textbf{\num{0.620 \pm 0.058}} & \num{0.654 \pm 0.043} & \textbf{\num{0.695 \pm 0.046}} & \num{0.718 \pm 0.049} & \textbf{\num{0.722 \pm 0.046}} \\
Expert & \num{0.573 \pm 0.017} & \num{0.646 \pm 0.010} & \num{0.645 \pm 0.002} & \num{0.653 \pm 0.009} & \num{0.637 \pm 0.023} \\
Prior & \num{0.581 \pm 0.007} & \textbf{\num{0.667 \pm 0.000}} & \num{0.665 \pm 0.000} & \textbf{\num{0.734 \pm 0.006}} & \num{0.661 \pm 0.027} \\
\hlineB{2}
Few-shot (16) + expert & \num{0.590 \pm 0.058} & \num{0.700 \pm 0.036} & \num{0.719 \pm 0.038} & \num{0.756 \pm 0.042} & \num{0.717 \pm 0.040} \\
Few-shot (16) + prior & \num{0.644 \pm 0.042} & \num{0.772 \pm 0.041} & \num{0.762 \pm 0.040} & \num{0.795 \pm 0.031} & \num{0.764 \pm 0.041} \\
Expert + prior & \underline{\num{0.732 \pm 0.006}} & \num{0.825 \pm 0.011} & \underline{\num{0.862 \pm 0.002}} & \num{0.779 \pm 0.005} & \num{0.764 \pm 0.027} \\
Few-shot (16) + expert + prior & \num{0.714 \pm 0.060} & \underline{\num{0.829 \pm 0.028}} & \num{0.843 \pm 0.029} & \underline{\num{0.818 \pm 0.028}} & \underline{\num{0.771 \pm 0.024}} \\
\hlineB{2}
TabPFN-3 Plus (16) & \num{0.719 \pm 0.058} & \num{0.719 \pm 0.058} & \num{0.719 \pm 0.058} & \num{0.723 \pm 0.064} & \num{0.719 \pm 0.058} \\
\bottomrule
\end{tabular}
}%
\caption{Comparison of ROC--AUC between prompting regimes with three in-context learning entities (\emph{few-shot} demonstrations, \emph{prior} natural-language task description, and \emph{expert} LLM-distilled decision rules) on the LLM-synthetic suite. Best single-entity regime per model in \textbf{bold}; best overall configuration per model \underline{underlined}.}
\label{tab:h3_icl_combinations}
\end{table*}

\textbf{Results}
Alone, \emph{Expert} never wins: it scores $0.573$--$0.653$ across the four
models (Table~\ref{tab:h3_icl_combinations}), below \emph{Prior} alone
($0.581$--$0.734$) and below \emph{Few-shot (16)} alone
($0.620$--$0.718$). Pairing \emph{Expert} with \emph{Prior} jumps to
$0.732$--$0.862$ ($+0.13$--$0.22$ over \emph{Expert} alone), clearing the
TabPFN-3 Plus baseline for three of four models; \emph{Few-shot + expert}
without \emph{prior} still sits below \emph{Few-shot} alone for the
smallest model ($0.590$ vs.\ $0.620$), while \emph{Few-shot + expert +
prior} reaches $0.714$--$0.843$. The advantage of having rules at all
shrinks as shots grow: by 16 examples, zero-shot \emph{Expert + prior}
already edges out \emph{Few-shot + expert + prior} for two of four
models -- the same budget identified in \qtitle{H2b} for saturation.

\textbf{Takeaway} Expert rules do not independently substitute for labeled data -- they
only help once paired with a contextual prior, and we have only verified
this for rules consistent with the model's own knowledge. Where they do
help, they compensate for missing demonstrations up to about 16 real
examples; beyond that, the marginal benefit of rules is no longer observed.

\section{Conclusion}
\label{sec:conclusion}

In this paper, we introduce TabLens, a benchmark for evaluating whether LLM-based in-context learning can substitute for labeled data in tabular classification under data scarcity. Our results show that \textbf{\qtitle{Q1}} zero-shot LLMs can replace limited data collection below a measurable complexity threshold, particularly for text-rich datasets; \textbf{\qtitle{Q2}} complete semantic context can replace additional demonstrations up to approximately 16 shots; and \textbf{\qtitle{Q3}} expert rules are beneficial only when paired with a matching contextual prior and within the same low-shot regime. The main limitation is the benchmark's focus on binary classification and a finite selection of models and datasets, which may restrict generalization to other tasks, domains, and adaptation settings.

\textbf{GenAI usage disclosure.} Generative AI tools were used to support the literature search, generate the synthetic datasets as described in the paper, and polish the manuscript's wording; all experimental design, analysis, and conclusions were developed and verified by the authors.

\bibliographystyle{iclr2026_conference}
\bibliography{references,software}

@article{wang2025towards,
  title={Towards data-centric ai: A comprehensive survey of traditional, reinforcement, and generative approaches for tabular data transformation},
  author={Wang, Dongjie and Huang, Yanyong and Ying, Wangyang and Bai, Haoyue and Gong, Nanxu and Wang, Xinyuan and Dong, Sixun and Zhe, Tao and Liu, Kunpeng and Xiao, Meng and others},
  journal={arXiv preprint arXiv:2501.10555},
  year={2025}
}

@article{chen2016xgboost,
  title={XGBoost: A Scalable Tree Boosting System},
  author={Chen, Tianqi},
  journal={Cornell University},
  year={2016}
}

@article{prokhorenkova2018catboost,
  title={CatBoost: unbiased boosting with categorical features},
  author={Prokhorenkova, Liudmila and Gusev, Gleb and Vorobev, Aleksandr and Dorogush, Anna Veronika and Gulin, Andrey},
  journal={Advances in neural information processing systems},
  volume={31},
  year={2018}
}

@article{ke2017lightgbm,
  title={Lightgbm: A highly efficient gradient boosting decision tree},
  author={Ke, Guolin and Meng, Qi and Finley, Thomas and Wang, Taifeng and Chen, Wei and Ma, Weidong and Ye, Qiwei and Liu, Tie-Yan},
  journal={Advances in neural information processing systems},
  volume={30},
  year={2017}
}

@inproceedings{arik2021tabnet,
  title={Tabnet: Attentive interpretable tabular learning},
  author={Arik, Sercan {\"O} and Pfister, Tomas},
  booktitle={Proceedings of the AAAI conference on artificial intelligence},
  volume={35},
  number={8},
  pages={6679--6687},
  year={2021}
}

@article{hollmann2022tabpfn,
  title={Tabpfn: A transformer that solves small tabular classification problems in a second},
  author={Hollmann, Noah and M{\"u}ller, Samuel and Eggensperger, Katharina and Hutter, Frank},
  journal={arXiv preprint arXiv:2207.01848},
  year={2022}
}

@inproceedings{hegselmann2023tabllm,
  title={Tabllm: Few-shot classification of tabular data with large language models},
  author={Hegselmann, Stefan and Buendia, Alejandro and Lang, Hunter and Agrawal, Monica and Jiang, Xiaoyi and Sontag, David},
  booktitle={International conference on artificial intelligence and statistics},
  pages={5549--5581},
  year={2023},
  organization={PMLR}
}

@article{slack2023tablet,
  title={Tablet: Learning from instructions for tabular data},
  author={Slack, Dylan and Singh, Sameer},
  journal={arXiv preprint arXiv:2304.13188},
  year={2023}
}

@article{manikandan2023language,
  title={Language models are weak learners},
  author={Manikandan, Hariharan and Jiang, Yiding and Kolter, J Zico},
  journal={Advances in Neural Information Processing Systems},
  volume={36},
  pages={50907--50931},
  year={2023}
}

@article{jaitly2023towards,
  title={Towards better serialization of tabular data for few-shot classification with large language models},
  author={Jaitly, Sukriti and Shah, Tanay and Shugani, Ashish and Grewal, Razik Singh},
  journal={arXiv preprint arXiv:2312.12464},
  year={2023}
}

@article{gardner2406large,
  title={Large scale transfer learning for tabular data via language modeling, 2024},
  author={Gardner, Josh and Perdomo, Juan C and Schmidt, Ludwig},
  journal={URL https://arxiv. org/abs/2406.12031}
}

@article{huang2012tabtransformer,
  title={Tabtransformer: tabular data modeling using contextual embeddings (2020)},
  author={Huang, Xin and Khetan, Ashish and Cvitkovic, Milan and Karnin, Zohar},
  journal={arXiv preprint arXiv:2012.06678},
  year={2012}
}

@article{gorishniy2410tabm,
  title={TabM: Advancing tabular deep learning with parameter-efficient ensembling, 2025},
  author={Gorishniy, Yury and Kotelnikov, Akim and Babenko, Artem},
  journal={URL https://arxiv. org/abs/2410.24210}
}

@inproceedings{chen2022danets,
  title={Danets: Deep abstract networks for tabular data classification and regression},
  author={Chen, Jintai and Liao, Kuanlun and Wan, Yao and Chen, Danny Z and Wu, Jian},
  booktitle={Proceedings of the AAAI Conference on Artificial Intelligence},
  volume={36},
  number={4},
  pages={3930--3938},
  year={2022}
}

@inproceedings{
    qu2025tabicl,
    title={Tab{ICL}: A Tabular Foundation Model for In-Context Learning on Large Data},
    author={Jingang QU and David Holzm{\"u}ller and Ga{\"e}l Varoquaux and Marine Le Morvan},
    booktitle={Forty-second International Conference on Machine Learning},
    year={2025},
    url={https://openreview.net/forum?id=0VvD1PmNzM}
}

@article{bischl2021openml,
  title={Openml: A benchmarking layer on top of openml to quickly create, download, and share systematic benchmarks},
  author={Bischl, Bernd and Casalicchio, Giuseppe and Feurer, Matthias and Gijsbers, Pieter and Hutter, Frank and Lang, Michel and Mantovani, Rafael Gomes and van Rijn, Jan N and Vanschoren, Joaquin},
  journal={NeurIPS--Track on Datasets and Benchmarks},
  year={2021}
}

@article{yang2025qwen3,
  title={Qwen3 technical report},
  author={Yang, An and Li, Anfeng and Yang, Baosong and Zhang, Beichen and Hui, Binyuan and Zheng, Bo and Yu, Bowen and Gao, Chang and Huang, Chengen and Lv, Chenxu and others},
  journal={arXiv preprint arXiv:2505.09388},
  year={2025}
}

@article{delong1988comparing,
  title={Comparing the areas under two or more correlated receiver operating characteristic curves: a nonparametric approach},
  author={DeLong, Elizabeth R and DeLong, David M and Clarke-Pearson, Daniel L},
  journal={Biometrics},
  pages={837--845},
  year={1988},
  publisher={JSTOR}
}

@article{sui2023gpt4table,
  title={Gpt4table: Can large language models understand structured table data? a benchmark and empirical study},
  author={Sui, Yuan and Zhou, Mengyu and Zhou, Mingjie and Han, Shi and Zhang, Dongmei},
  journal={arXiv preprint ArXiv:2305.13062},
  year={2023}
}

@inproceedings{wu2025tablebench,
  title={Tablebench: A comprehensive and complex benchmark for table question answering},
  author={Wu, Xianjie and Yang, Jian and Chai, Linzheng and Zhang, Ge and Liu, Jiaheng and Du, Xeron and Liang, Di and Shu, Daixin and Cheng, Xianfu and Sun, Tianzhen and others},
  booktitle={Proceedings of the AAAI Conference on Artificial Intelligence},
  volume={39},
  number={24},
  pages={25497--25506},
  year={2025}
}

@article{erickson2025tabarena,
  title={Tabarena: A living benchmark for machine learning on tabular data},
  author={Erickson, Nick and Purucker, Lennart and Tschalzev, Andrej and Holzm{\"u}ller, David and Desai, Prateek Mutalik and Salinas, David and Hutter, Frank},
  journal={arXiv preprint arXiv:2506.16791},
  year={2025}
}

@article{mcelfresh2023neural,
  title={When do neural nets outperform boosted trees on tabular data?},
  author={McElfresh, Duncan and Khandagale, Sujay and Valverde, Jonathan and Prasad C, Vishak and Ramakrishnan, Ganesh and Goldblum, Micah and White, Colin},
  journal={Advances in Neural Information Processing Systems},
  volume={36},
  pages={76336--76369},
  year={2023}
}

@article{bac2021scikit,
  title={Scikit-dimension: a python package for intrinsic dimension estimation},
  author={Bac, Jonathan and Mirkes, Evgeny M and Gorban, Alexander N and Tyukin, Ivan and Zinovyev, Andrei},
  journal={Entropy},
  volume={23},
  number={10},
  pages={1368},
  year={2021},
  publisher={MDPI}
}

@article{fukunaga1971algorithm,
  title={An algorithm for finding intrinsic dimensionality of data},
  author={Fukunaga, Keinosuke and Olsen, David R},
  journal={IEEE Transactions on computers},
  volume={100},
  number={2},
  pages={176--183},
  year={1971},
  publisher={IEEE}
}

@article{mirkes2020fractional,
  title={Fractional norms and quasinorms do not help to overcome the curse of dimensionality},
  author={Mirkes, Evgeny M and Allohibi, Jeza and Gorban, Alexander},
  journal={Entropy},
  volume={22},
  number={10},
  pages={1105},
  year={2020},
  publisher={MDPI}
}

@article{fan2010intrinsic,
  title={Intrinsic dimension estimation of data by principal component analysis},
  author={Fan, Mingyu and Gu, Nannan and Qiao, Hong and Zhang, Bo},
  journal={arXiv preprint arXiv:1002.2050},
  year={2010}
}

@article{levina2004maximum,
  title={Maximum likelihood estimation of intrinsic dimension},
  author={Levina, Elizaveta and Bickel, Peter},
  journal={Advances in neural information processing systems},
  volume={17},
  year={2004}
}

@article{facco2017estimating,
  title={Estimating the intrinsic dimension of datasets by a minimal neighborhood information},
  author={Facco, Elena and d’Errico, Maria and Rodriguez, Alex and Laio, Alessandro},
  journal={Scientific reports},
  volume={7},
  number={1},
  pages={12140},
  year={2017},
  publisher={Nature Publishing Group UK London}
}

@article{carter2009local,
  title={On local intrinsic dimension estimation and its applications},
  author={Carter, Kevin M and Raich, Raviv and Hero III, Alfred O},
  journal={IEEE Transactions on Signal Processing},
  volume={58},
  number={2},
  pages={650--663},
  year={2009},
  publisher={IEEE}
}

@article{gorla2026illusion,
  title={The Illusion of Generalization: Re-examining Tabular Language Model Evaluation},
  author={Gorla, Aditya and Puduppully, Ratish},
  journal={arXiv preprint arXiv:2602.04031},
  year={2026}
}

@article{grinsztajn2026tabpfn3,
  title  = {TabPFN-3: Technical Report},
  author = {Grinsztajn, L{\'e}o and others},
  journal= {arXiv preprint arXiv:2605.13986},
  year   = {2026},
  url    = {https://arxiv.org/abs/2605.13986}
}

@inproceedings{han2024featllm,
  title     = {Large Language Models Can Automatically Engineer Features for Few-Shot Tabular Learning},
  author    = {Han, Sungwon and Yoon, Jinsung and Arik, Sercan {\"O}. and Pfister, Tomas},
  booktitle = {International Conference on Machine Learning (ICML)},
  year      = {2024}
}

@inproceedings{knauer2025llmtrees,
  title     = {{``Oh LLM, I'm Asking Thee, Please Give Me a Decision Tree'': Zero-Shot Decision Tree Induction and Embedding with Large Language Models}},
  author    = {Knauer, Ricardo and Koddenbrock, Mario and Wallsberger, Raphael and Brisson, Nicholas M. and Duda, Georg N. and Falla, Deborah and Evans, David W. and Rodner, Erik},
  booktitle = {Proceedings of the 31st ACM SIGKDD Conference on Knowledge Discovery and Data Mining (KDD)},
  year      = {2025}
}

@article{gardner2024large,
  title={Large scale transfer learning for tabular data via language modeling},
  author={Gardner, Josh and Perdomo, Juan C and Schmidt, Ludwig},
  journal={Advances in Neural Information Processing Systems},
  volume={37},
  pages={45155--45205},
  year={2024}
}

@article{silvestri2025evaluating,
  title={Evaluating latent knowledge of public tabular datasets in large language models},
  author={Silvestri, Matteo and Veglianti, Fabiano and Giorgi, Flavio and Silvestri, Fabrizio and Tolomei, Gabriele},
  journal={arXiv preprint arXiv:2510.20351},
  year={2025}
}

@article{bordt2024elephants,
  title={Elephants never forget: Memorization and learning of tabular data in large language models},
  author={Bordt, Sebastian and Nori, Harsha and Rodrigues, Vanessa and Nushi, Besmira and Caruana, Rich},
  journal={arXiv preprint arXiv:2404.06209},
  year={2024}
}

@article{shi2021benchmarking,
  title={Benchmarking multimodal automl for tabular data with text fields},
  author={Shi, Xingjian and Mueller, Jonas and Erickson, Nick and Li, Mu and Smola, Alexander J},
  journal={arXiv preprint arXiv:2111.02705},
  year={2021}
}

@article{mraz2025towards,
  title={Towards benchmarking foundation models for tabular data with text},
  author={Mr{\'a}z, Martin and Das, Breenda and Gupta, Anshul and Purucker, Lennart and Hutter, Frank},
  journal={arXiv preprint arXiv:2507.07829},
  year={2025}
}

@article{arazi2026multabench,
  title={MulTabench: Benchmarking multimodal tabular learning with text and image},
  author={Arazi, Alan and Shapira, Eilam and Grunblat, Shoham and Ventura, Mor and Hoffer, Elad and Blayer, Gioia and Holzm{\"u}ller, David and Purucker, Lennart and Varoquaux, Ga{\"e}l and Hutter, Frank},
  journal={arXiv preprint arXiv:2605.10616},
  year={2026}
}

@article{blayer2026strable,
  title={StrTable: Benchmarking tabular machine learning with strings},
  author={Blayer, Gioia and Kim, Myung Jun and Lefebvre, F{\'e}lix and Purucker, Lennart and Arazi, Alan and Shapira, Eilam and Reichart, Roi and Hutter, Frank and Le Morvan, Marine and Holzm{\"u}ller, David},
  journal={arXiv preprint arXiv:2605.12292},
  year={2026}
}

@misc{priorlabs2025tabpfn3,
  title={{TabPFN}-3 (Plus): Prior-data fitted networks for tabular classification},
  author={{Prior Labs}},
  year={2025},
  howpublished={\url{https://priorlabs.ai/tabpfn-3}},
  note={Accessed 2026}
}

@inproceedings{xie2022explanation,
  title={An Explanation of In-Context Learning as Implicit {B}ayesian Inference},
  author={Xie, Sang Michael and Raghunathan, Aditi and Liang, Percy and Ma, Tengyu},
  booktitle={International Conference on Learning Representations},
  year={2022}
}

@inproceedings{cheng2025tabfsbench,
  title={{TabFSBench}: Tabular Benchmark for Feature Shifts in Open Environments},
  author={Cheng, Zi-Jian and Jia, Zi-Yi and Zhou, Zhi and Guo, Lan-Zhe and Li, Yu-Feng},
  booktitle={International Conference on Machine Learning (ICML)},
  year={2025}
}

@article{demsar2006statistical,
  title={Statistical comparisons of classifiers over multiple data sets},
  author={Dem{\v{s}}ar, Janez},
  journal={Journal of Machine Learning Research},
  volume={7},
  pages={1--30},
  year={2006}
}

@article{holm1979simple,
  title={A simple sequentially rejective multiple test procedure},
  author={Holm, Sture},
  journal={Scandinavian Journal of Statistics},
  volume={6},
  number={2},
  pages={65--70},
  year={1979},
  publisher={JSTOR}
}

@article{fawcett2006introduction,
  title={An introduction to {ROC} analysis},
  author={Fawcett, Tom},
  journal={Pattern Recognition Letters},
  volume={27},
  number={8},
  pages={861--874},
  year={2006},
  publisher={Elsevier}
}

% ===== Appendix =====
\newpage
\appendix
% =============================================================================
% APPENDIX — STRUCTURED TEMPLATE (v3 — two-column-friendly)
% =============================================================================
% Organization:
%   A. Datasets   (expanded, two-column-friendly with table*)
%   B. Prompts
%   C. Statistical methodology
%   D. RQ1 — zero-shot competitiveness
%   E. RQ2 — in-context vs prior
%   F. RQ3 — few-shot effectiveness
%   G. RQ4 — removing in-context info in few-shot
%   H. RQ5 — combining all three sources
%   I. RQ6 — dataset complexity
%   J. Per-domain raw tables (shots & serializations, 9 domains each)
%   K. Additional figures
%
% Two-column compatibility notes:
%   - All wide tables use the table*/figure* environments so they span both
%     columns. Narrow tables stay in table/figure.
%   - The dataset table is wrapped in \begin{table*}[t!] ... \end{table*} and
%     uses \resizebox to fit textwidth.
%   - Cells marked TODO are values I am not certain about — fill in or remove.
% =============================================================================

\newpage
% Appendix inherits the document's two-column layout. Wide tables and
% figures use the starred (table*/figure*) environments to span both columns.
\appendix

% =============================================================================
\section{Key Result Figures}
\label{app:key_figures}
% =============================================================================

This section collects the two headline figures referenced most often from
the main text -- the per-domain zero-shot radar (H1a) and the combined
GPT-4o-mini vs.\ TabPFN-3 Plus text-tabular comparison (H1c) -- in one
place, ahead of the per-question appendix sections that follow.

\begin{figure*}[t]
    \centering
    \includegraphics[width=0.6\textwidth]{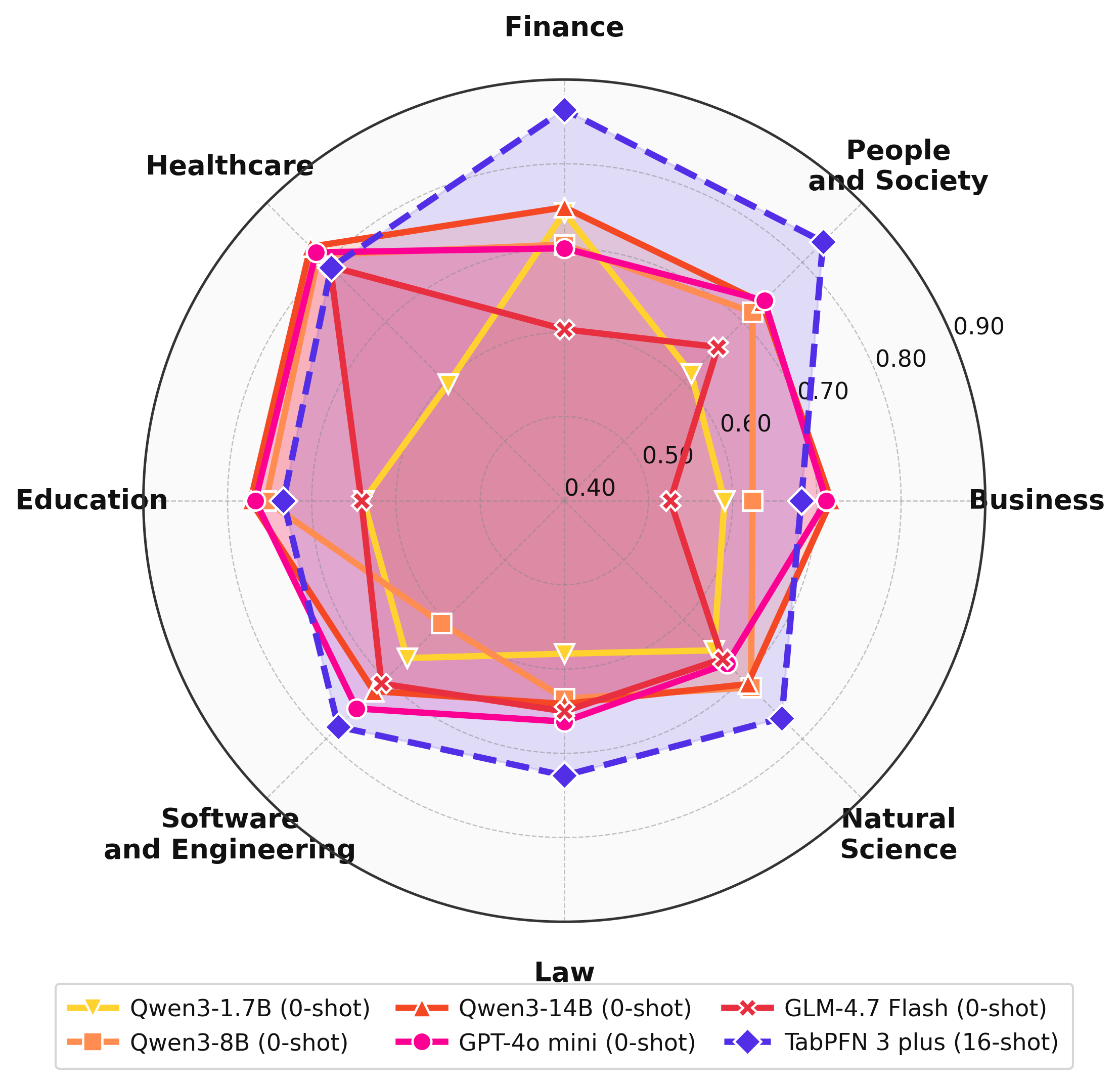}
    \caption{Per-domain mean ROC--AUC at 0-shot for the five LLMs vs.\
    16-shot TabPFN-3 Plus (dashed reference), on the 24 core real-world
    datasets grouped into 8 knowledge domains. Referenced from
    Section~\ref{sec:q1_h1a} (H1a).}
    \label{fig:rq1_radar}
    \Description{Per-domain zero-shot LLM performance compared with 16-shot TabPFN-3 Plus across eight knowledge domains.}
\end{figure*}

\begin{figure*}[t]
    \centering
    \includegraphics[width=0.85\textwidth]{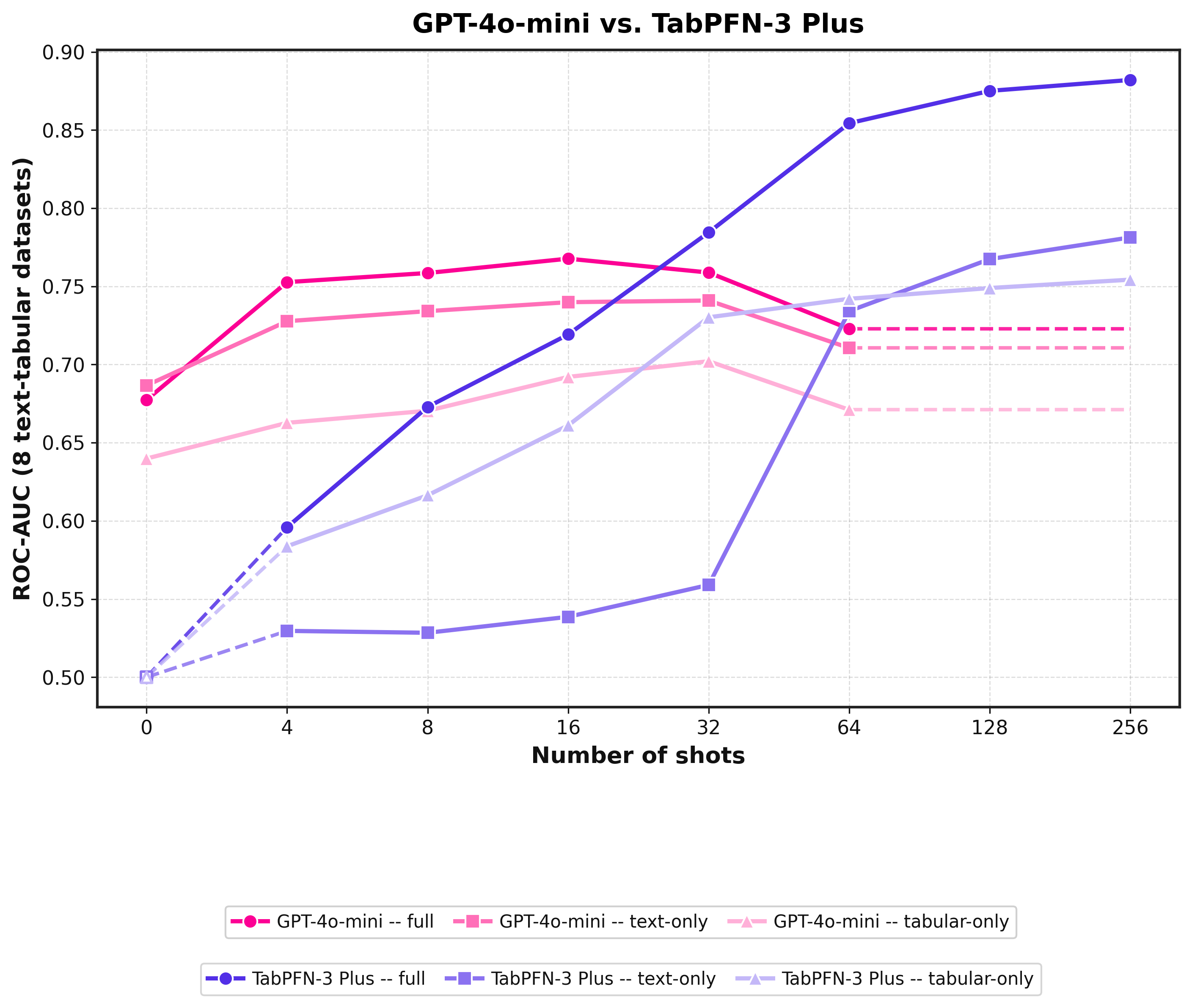}
    \caption{GPT-4o-mini vs.\ TabPFN-3 Plus across shot counts on the 8
    text-tabular datasets, for the \emph{full}, \emph{text-only}, and
    \emph{tabular-only} column variants. Every solid segment is a real
    measurement; TabPFN-3 Plus has no true zero-shot mode, so its curve is
    extended left to chance level (dashed) at $k=0$, and GPT-4o-mini's
    last measured point (64 shots) is extended flat and dashed out to
    TabPFN-3 Plus's higher shot counts purely to keep both curves on a
    shared axis. Referenced from Section~\ref{sec:q1_text} (H1c).}
    \label{fig:texttab_niche}
    \Description{GPT-4o-mini and TabPFN-3 Plus performance across shot counts for full, text-only, and tabular-only inputs on the 8 text-tabular datasets.}
\end{figure*}

% =============================================================================
\section{Detailed evaluation protocol}
\label{app:protocol}
% =============================================================================

This section states the task construction and evaluation protocol of each
research question in full, expanding the condensed descriptions given in the
main text.

\subsection{RQ1: How competitive are LLMs in zero-shot tabular classification?}

\textbf{Setup}
This benchmark evaluates the zero-shot performance of multiple large language models (LLMs) on a binary classification task. 

\textbf{Experimental Design}
Building on this setup, we evaluate 5 LLMs, 2 inference modes, and 5 serialization regimes as mentioned in Section~\ref{sec:3}. For this RQ, 24 binary classification datasets from the core real-world suite are used (see Table~\ref{tab:llmtabbench_suites}). 

\textbf{Statistical Analysis}
To determine whether zero-shot performance is statistically superior to
random guessing, we treat the 24 classic datasets as the independent
experimental units. For each model, the dataset-level mean ROC--AUC is compared
with $0.5$ using a one-sided Wilcoxon signed-rank test; Holm-adjusted
$p$-values control family-wise error across the five models. Full details are
given in Appendix~\ref{app:chance_test}.

\subsection{RQ2: To what extent does zero-shot performance depend on in-context information versus LLM prior knowledge?}

\textbf{Setup} This setting assesses whether zero-shot performance is primarily attributable to pretrained knowledge or to in-context information by measuring performance degradation under progressive removal of in-context information.

\textbf{Experimental design} We use the same fixed grid as above, except for additional ablations explained below.

\begin{itemize}
    \item \textbf{Context Prompt (2 Levels):}
    \begin{itemize}
        \item \textbf{With Context:} A prompt containing an explicit task description specifying the domain, classification objective, and target labels. 
        \item \textbf{Without Context:} A neutral instruction that does not disclose the domain, task, or target values (e.g., \textit{"Perform binary classification for the following row."}). 
    \end{itemize}

    \item \textbf{Headers (2 Levels):}
    \begin{itemize}
        \item \textbf{Semantic:} Real columns' names. 
        \item \textbf{Generic:} Placeholders instead of them. 
    \end{itemize}
    
    \item \textbf{Graded Context Configuration (4 Levels):} combining the Context Prompt (\textit{With Context} / \textit{Without Context}) with the Headers (\textit{Semantic} / \textit{Generic}) types (see Table~\ref{tab:prior_knowledge_gradation} for description).
\end{itemize}

\begin{table}[tb]
\centering
\caption{Descriptions of levels in the Graded Context Configuration setup
referenced from Section~\ref{sec:q2_h2a}.}
\label{tab:prior_knowledge_gradation}
\renewcommand{\arraystretch}{1.25}
\begin{tabularx}{\linewidth}{ >{\centering\arraybackslash}X >{\centering\arraybackslash}X >{\centering\arraybackslash}X >{\centering\arraybackslash}X }
\toprule
\textbf{Configuration Name} & \textbf{Configuration Index} & \textbf{Names of Columns} & \textbf{Prompt}  \\
\midrule
Maximal Context & \textbf{Cols-Context} & Semantic & Context \\
Semantic Headers Only & \textbf{Cols-NoContext} & Semantic & No-context  \\
Context Prompt Only & \textbf{NoCols-Context} & Generic & Context \\
Minimal Context & \textbf{NoCols-NoContext} & Generic & No-context  \\
\bottomrule
\end{tabularx}
\end{table}

\subsection{RQ4: How does removing in-context information affect few-shot performance across shot regimes?}

\noindent\textbf{Setup} This section examines how few-shot examples interact with explicit contextual information in binary classification, specifically whether they enable models to compensate for missing task descriptions or semantic feature names.

\textbf{Experimental Design}
This study extends the design established in RQ2 by introducing few-shot examples while maintaining the same systematic variation of contextual information.

\textbf{Procedure}

For each unique combination of factors (Model $\times$ Serialization $\times$ Context Prompt $\times$ Graded Configuration $\times$ $k$), $k$ correctly labeled examples are sampled from the training set with a fixed seed. A prompt is then constructed by including either a context-specific prompt or a neutral instruction, appending the $k$ few-shot examples formatted according to the chosen serialization method, and adding the test query using the same format. 

To evaluate whether few-shot examples can compensate for missing contextual information, we conduct planned contrasts comparing minimal-context conditions with a high number of few-shot examples (e.g., ($k$ = 64)) to maximal-context conditions with a small number of examples (e.g., ($k$ = 4)). In addition, we analyze how performance scales with increasing $k$ across different contextual configurations.

\subsection{RQ5: Does combining in-context information, few-shot examples, and external instructions improve performance?}

\noindent\textbf{Setup} This setting investigates whether combining explicit in-context information, few-shot demonstrations, and external knowledge in the form of decision rules improves predictive performance beyond using each source in isolation.

\textbf{Experimental Design}
This study extends the design from RQ4 by introducing two novel experimental dimensions: synthetic datasets containing model-specific prior knowledge, and explicit decision rules as external instructions.

The experimental design varies two additional factors. First, we consider two dataset types: real datasets, consisting of the 24 datasets used in previous RQs, and synthetic datasets, which are model-generated counterparts created by prompting the same model to produce synthetic examples reflecting its learned priors about the domain, task, and label distribution (see Appendix~\ref{app:prompts} for the corresponding prompt). 
Second, we vary the type of external instructions provided in the prompt across three conditions: a baseline condition with no external instructions beyond context and few-shot examples; a condition where synthetic dataset examples generated by the model are included alongside or in place of real few-shot examples; and a condition where the prompt includes explicit decision rules generated by the LLM (e.g., "If $x_1 > 0.5$ and $x_3 < 2.0$, then class = 1") and expressed as if–then statements.

\subsection{RQ6: How does dataset complexity influence LLM performance on tabular tasks?}

\textbf{Experiment Design}
We use the same models, inference modes, and zero-/few-shot settings as in the previous RQs.
Data are serialized using two masked formats that are functionally equivalent in this context-free setting, given the absence of semantic feature headers. A single context-free prompt is employed, containing either only the formatted query instance in the zero-shot setting or $k$ demonstration examples followed by the query instance in the few-shot setting. For each of the 35 synthetic datasets, experiments are conducted across all relevant factor combinations.

\paragraph{Statistical Analysis}
To assess whether LLMs can learn classification rules from data alone without semantic context, we compare $k$-shot performance to a zero-shot baseline using the same procedure as in RQ3. The null hypothesis states that $k$-shot performance with pure data patterns does not exceed zero-shot performance across all datasets.

% =============================================================================
\section{Datasets}
\label{app:datasets}
% =============================================================================

We use four dataset suites:
\textbf{(i)} a \emph{core real-world suite} of 24 binary-classification tasks
spanning eight knowledge domains (Section~\ref{app:real_datasets});
\textbf{(ii)} a \emph{new suite} of 8 datasets, one per domain, first
released \emph{after} every evaluated LLM's knowledge cutoff (October 2024),
used to probe data leakage and out-of-cutoff generalisation
(Section~\ref{app:temporal_datasets});
\textbf{(iii)} a \emph{synthetic suite} generated from MLP priors at seven
controlled difficulty levels (Section~\ref{app:synthetic_datasets});
\textbf{(iv)} an \emph{LLM-synthetic suite} obtained by prompting Qwen3-8B or
GPT-4o-mini to fabricate tabular records matching the schemas of the 24 core
datasets (Section~\ref{app:llm_synthetic_datasets}).
Tables~\ref{tab:datasets_real},~\ref{tab:datasets_temporal} below list
per-dataset metadata. Imbalance (\% positive) refers to the prevalence of
the positive class in the full dataset.

\begin{table*}[t!]
\centering
\caption{Core real-world suite (24 binary classification datasets). \texttt{ID} = identifier used in all per-dataset tables throughout the paper. \#Feat. = total input features (target excluded); the sub-columns give the numerical/categorical split. \% Pos.\ = prevalence of the positive class. ${}^{\ast}$Dataset was class-stratified subsampled to 1{,}000 instances with seed \texttt{42}; original sizes are listed in the running text. ${}^{\ddagger}$\texttt{irish}: ownership retained by Greaney \& Kelleghan; usable only as an example for the development of statistical methods.}
\label{tab:datasets_real}
\scriptsize
\setlength{\tabcolsep}{3pt}
\renewcommand{\arraystretch}{1.1}
\resizebox{\textwidth}{!}{%
\begin{tabular}{@{}llrrrrlrlp{4.2cm}@{}}
\toprule
\textbf{ID} & \textbf{Source} & \textbf{Size} & \textbf{\#Feat.} & \textbf{Num.} & \textbf{Cat.} & \textbf{Target} & \textbf{\% Pos.} & \textbf{License} & \textbf{Domain / Task (target = 1 means \ldots)} \\
\midrule
\multicolumn{10}{l}{\textit{Healthcare}} \\
\midrule
diabetes &
\href{https://www.openml.org/d/37}{OpenML-37} &
768 & 8 & 8 & 0 & Outcome & 34.9 & CC0 &
Pima Indians diabetes diagnosis from physiological measurements. \emph{Target = 1}: patient has diabetes. \\
transfusion &
\href{https://www.openml.org/d/1464}{OpenML-1464} &
748 & 4 & 4 & 0 & Donated 2007 & 23.8 & CC BY 4.0 &
Blood-donor RFM-style features (Hsinchu, Taiwan). \emph{Target = 1}: donor gave blood in March 2007. \\
cancer &
\href{https://www.openml.org/d/15}{OpenML-15} &
699 & 9 & 9 & 0 & Class & 34.5 & CC BY 4.0 &
Wisconsin breast-cancer cell-nucleus cytology. \emph{Target = 1}: cell is malignant; 0: benign. \\
\midrule
\multicolumn{10}{l}{\textit{Business}} \\
\midrule
marketing &
\href{https://www.openml.org/d/46940}{OpenML-46940} &
2240 & 25 & 16 & 9 & Response & 14.9 & CC0 &
Customer demographic and transactional features. \emph{Target = 1}: customer responded to the marketing campaign. \\
telco &
\href{https://www.openml.org/d/42178}{OpenML-42178} &
1000$^{\ast}$ & 19 & 3 & 16 & Churn & 26.5 & IBM &
Telecom customer contract, billing, and usage features. \emph{Target = 1}: customer churned (left within the last month). \\
hiring &
\href{https://www.kaggle.com/datasets/avikumart/hrdatasetclassif}{Kaggle (avikumart)} &
1000$^{\ast}$ & 17 & 8 & 9 & label & 81.4 & Kaggle terms &
HR analytics features for job candidates (interview, offer, hiring pipeline). \emph{Target = 1}: candidate joined the company; 0: did not join. \\
\midrule
\multicolumn{10}{l}{\textit{People \& Society}} \\
\midrule
adult &
\href{https://www.kaggle.com/datasets/uciml/adult-census-income}{UCI Adult} &
1000$^{\ast}$ & 13 & 6 & 8 & income\,$>$50K & 23.9 & CC BY 4.0 &
US 1994 census demographic and employment features. \emph{Target = 1}: annual income $>$\$50K. \\
fitness &
\href{https://www.openml.org/d/46927}{OpenML-46927} &
1500 & 6 & 3 & 3 & attended & 30.2 & Public Domain &
GoalZone fitness-club member and class registration (DataCamp, curated by TabArena). \emph{Target = 1}: member attended the booked class. \\
tech\_mental\_health\_survey &
\href{https://www.kaggle.com/datasets/osmi/mental-health-in-tech-survey}{Kaggle (OSMI)} &
51 & 22 & 1 & 21 & label & 56.9 & CC BY-SA &
2014 OSMI survey of tech-industry employees (subsampled version). \emph{Target = 1}: respondent feels employer takes mental health as seriously as physical health. \\
\midrule
\multicolumn{10}{l}{\textit{Finance}} \\
\midrule
credit &
\href{https://www.openml.org/d/31}{OpenML-31} &
1000 & 20 & 7 & 13 & class & 30.0 & CC BY 4.0 &
German Credit (Statlog) bank-client financial and personal attributes. \emph{Target = 1}: ``good'' credit risk (likely to repay); 0: ``bad'' risk. \\
audit &
\href{https://www.kaggle.com/datasets/sid321axn/audit-data}{UCI / Kaggle} &
776 & 26 & 23 & 3 & label & 39.3 & CC BY 4.0 &
Government-firm audit features (financial scores, risk scores, sector). \emph{Target = 1}: firm has fraud risk; 0: no fraud risk. \\
fraud &
\href{https://www.kaggle.com/datasets/neharoychoudhury/credit-card-fraud-data}{Kaggle (neharoychoudhury)} &
1000$^{\ast}$ & 14 & 6 & 8 & is\_fraud & 12.8 & Kaggle terms &
Credit-card transaction features (merchant, category, amount, location, time). \emph{Target = 1}: transaction is fraudulent; 0: legitimate. \\
\bottomrule
\end{tabular}%
}
\end{table*}

\begin{table*}[t!]
\centering
\caption{Core real-world suite (continued): the education, software engineering, crimes \& justice, and natural science domains. Column definitions as in Table~\ref{tab:datasets_real}.}
\label{tab:datasets_real_cont}
\scriptsize
\setlength{\tabcolsep}{3pt}
\renewcommand{\arraystretch}{1.1}
\resizebox{\textwidth}{!}{%
\begin{tabular}{@{}llrrrrlrlp{4.2cm}@{}}
\toprule
\textbf{ID} & \textbf{Source} & \textbf{Size} & \textbf{\#Feat.} & \textbf{Num.} & \textbf{Cat.} & \textbf{Target} & \textbf{\% Pos.} & \textbf{License} & \textbf{Domain / Task (target = 1 means \ldots)} \\
\midrule\multicolumn{10}{l}{\textit{Education}} \\
\midrule
tae &
\href{https://www.openml.org/d/955}{OpenML-955} &
151 & 5 & 3 & 2 & binaryClass & 34.0 & Public &
University TA evaluations (language, instructor, course, semester, group size). Binarised version: original 3-class target (low/medium/high) re-labelled with the majority class as positive (\texttt{P}) and the rest as negative (\texttt{N}). \\
irish &
\href{https://www.openml.org/d/451}{OpenML-451} &
500 & 5 & 2 & 3 & Leaving\_Cert & 44.4 & Research-only$^{\ddagger}$ &
Irish schoolchildren (age 11) educational-transition records (Greaney \& Kelleghan, 1984). \emph{Target = 1}: child took the Leaving Certificate; 0: did not. \\
student\_famsup &
\href{https://archive.ics.uci.edu/dataset/320/student+performance}{UCI-320} &
1298 & 32 & 4 & 28 & target & 61.3 & CC BY 4.0 &
Two Portuguese schools (math + Portuguese combined); features include school, demographics, study habits, and grades G1--G3. \emph{Target = 1}: student receives family educational support (\texttt{famsup}); 0: does not. \\
\midrule
\multicolumn{10}{l}{\textit{Software Engineering}} \\
\midrule
pc4 &
\href{https://www.openml.org/d/1049}{OpenML-1049} &
1458 & 37 & 37 & 0 & defects (c) & 12.2 & Public (NASA) &
Static code metrics of a NASA flight-software module. \emph{Target = 1}: module contains defects. \\
kc1 &
\href{https://www.openml.org/d/1067}{OpenML-1067} &
2109 & 21 & 21 & 0 & defects & 15.4 & Public (NASA) &
Static code metrics of a NASA storage-management module. \emph{Target = 1}: module contains defects. \\
steel &
\href{https://www.openml.org/d/1504}{OpenML-1504} &
1941 & 33 & 33 & 0 & Class & 65.3 & CC BY 4.0 &
Steel-plate fault sensor readings. \emph{Target = 1}: plate has defects other than the 6 named fault types; 0: no other defects. Originally 7 classes. \\
\midrule
\multicolumn{10}{l}{\textit{Crimes \& Justice}} \\
\midrule
compas &
\href{https://www.openml.org/d/42193}{OpenML-42193} &
1000$^{\ast}$ & 13 & 5 & 8 & two\_year\_recid & 45.5 & MIT &
ProPublica COMPAS criminal-defendant features (Broward County). \emph{Target = 1}: defendant re-offended within 2 years. \\
vote &
\href{https://www.openml.org/d/56}{OpenML-56} &
435 & 16 & 0 & 16 & Class & 61.4 & CC BY 4.0 &
1984 US Congressional voting records on 16 key votes. \emph{Target = 1}: representative is a Democrat; 0: Republican. \\
san\_francisco\_crimes &
\href{https://www.kaggle.com/datasets/kaggle/san-francisco-crime-classification}{Kaggle (SFPD)} &
1000$^{\ast}$ & 8 & 2 & 6 & label & 32.9 & Public (SFPD) &
SF Police incident records (date, district, address, coordinates, primary type). \emph{Target = 1}: incident was resolved with an arrest or citation; 0: not resolved with arrest. Originally $\sim$39 classes. \\
\midrule
\multicolumn{10}{l}{\textit{Natural Science}} \\
\midrule
biodegr &
\href{https://www.openml.org/d/1494}{OpenML-1494} &
1055 & 41 & 41 & 0 & Class & 33.7 & CC BY 4.0 &
41 molecular descriptors of chemical compounds. \emph{Target = 1}: ready-biodegradable; 0: not ready-biodegradable. \\
seismic\_bumps &
\href{https://www.kaggle.com/datasets/pranabroy94/seismic-bumps-data-set}{UCI / Kaggle} &
2584 & 18 & 14 & 4 & class & 6.6 & CC BY 4.0 &
Sensor readings from a Polish coal mine. \emph{Target = 1}: hazardous state — high-energy seismic bump will occur in the next shift; 0: non-hazardous. \\
bbbp &
\href{https://github.com/GLambard/Molecules_Dataset_Collection}{MoleculeNet} &
2039 & 18 & 18 & 0 & p\_np & 76.5 & MIT &
Precomputed molecular descriptors. \emph{Target = 1}: molecule penetrates the blood--brain barrier; 0: does not penetrate. \\
\bottomrule
\end{tabular}%
}
\end{table*}

\subsection{Real Datasets (core suite)}
\label{app:real_datasets}

The core suite covers eight knowledge domains: healthcare, business, people
\& society, finance, education, software engineering, crimes \& justice, and
natural science (three datasets per domain). Datasets were selected to span
small ($n=151$, \texttt{tae}) to large ($n\approx 49\text{k}$, \texttt{adult})
sample sizes, low ($d=4$) to high ($d=41$) dimensionality, and both balanced
($\sim$50\%) and severely imbalanced ($<10\%$) class distributions.
Three originally multi-class tasks (\texttt{tae}, \texttt{steel},
\texttt{san\_francisco\_crimes}) were collapsed to binary problems by
grouping the two most populous classes against the rest. The full
per-dataset metadata is given in Table~\ref{tab:datasets_real}.

\subsection{New Suite (post-cutoff datasets)}
\label{app:temporal_datasets}

To probe whether observed zero-shot performance reflects memorisation, we
assemble a parallel suite of eight datasets — one per knowledge domain —
first released \emph{after} the knowledge cutoff of every LLM evaluated
in this paper (October 2024). All datasets are binary-classification tasks
that follow the same preprocessing pipeline as the core suite.

\paragraph{Subsampling for large datasets.}
Several datasets in both the core and new suites contain more than
5{,}000 instances. To keep the experimental wall-clock and inference cost
tractable for closed-source and large open models, we subsample every
dataset whose original size exceeds 5{,}000 down to exactly 1{,}000
instances. Subsampling is class-stratified (so the positive-class rate
reported in Tables~\ref{tab:datasets_real} and~\ref{tab:datasets_temporal}
is preserved) and uses a fixed random seed of \texttt{42} for full
reproducibility. The \texttt{Size} columns in both tables therefore
report the size used in the experiments, not the source size. Affected
datasets are: in the core suite, \texttt{telco} (orig.\ 7{,}043 $\to$
1{,}000), \texttt{adult} (orig.\ 48{,}842 $\to$ 1{,}000),
\texttt{hiring} (orig.\ 8{,}995 $\to$ 1{,}000),
\texttt{fraud} (orig.\ 14{,}446 $\to$ 1{,}000),
\texttt{san\_francisco\_crimes} (orig.\ $\sim$878{,}049 $\to$ 1{,}000),
\texttt{compas} (orig.\ 5{,}278 $\to$ 1{,}000), and
\texttt{bbbp} (orig.\ 2{,}039, kept as is).
The new suite contains no dataset exceeding 5{,}000 instances.

\begin{table*}[t!]
\centering
\caption{New suite (post-cutoff): 8 datasets, one per knowledge domain, published after October 2024. \texttt{Size}, \texttt{\#Feat.}, \texttt{Num.}, \texttt{Cat.}, and \texttt{\% Pos.}\ are computed directly from the released CSV files (index columns excluded). The target variable is hand-constructed for each dataset; the semantics of the positive class are described in the \emph{Domain / Task} column. ${}^{\dagger}$Mendeley default licence (CC BY 4.0) — applies unless the dataset author explicitly opted for a different one.}
\label{tab:datasets_temporal}
\scriptsize
\setlength{\tabcolsep}{3pt}
\renewcommand{\arraystretch}{1.1}
\resizebox{\textwidth}{!}{%
\begin{tabular}{@{}llrrrrlrlp{4.2cm}@{}}
\toprule
\textbf{ID} & \textbf{Source} & \textbf{Size} & \textbf{\#Feat.} & \textbf{Num.} & \textbf{Cat.} & \textbf{Target} & \textbf{\% Pos.} & \textbf{License} & \textbf{Domain / Task (target = 1 means \ldots)} \\
\midrule
bank\_credit\_scoring &
\href{https://www.kaggle.com/datasets/kapturovalexander/bank-credit-risk-assessment}{Kaggle (kapturovalexander)} &
1000 & 15 & 4  & 11 & target & 42.8 & Kaggle terms &
Finance — bank credit risk assessment. \emph{Target = 1}: client has high credit scoring (credible, likely to repay the loan); 0: low credit scoring. \\
callcenter &
\href{https://data.mendeley.com/datasets/8854gk66td/2}{Mendeley 8854gk66td} &
1020 & 12 & 3  & 9  & target & 88.0 & CC BY 4.0$^{\dagger}$ &
Business — Kuwait 151 health hotline caller survey. \emph{Target = 1}: caller was satisfied with the call-centre service; 0: dissatisfied. \\
postpartum &
\href{https://data.mendeley.com/datasets/4nznnrk8cg/2}{Mendeley 4nznnrk8cg} &
800  & 44 & 1  & 43 & target & 43.8 & CC BY 4.0$^{\dagger}$ &
Healthcare — hospital female patients, postpartum depression. \emph{Target = 1}: patient has postpartum depression; 0: does not. \\
machine &
\href{https://www.kaggle.com/datasets/mujtabamatin/dataset-for-machine-failure-detection}{Kaggle (mujtabamatin)} &
1000 & 5  & 4  & 1  & target & 30.0 & Kaggle terms &
Software/Engineering — industrial machine sensor data. \emph{Target = 1}: machine is at risk of failure; 0: normal operation. \\
stars &
\href{https://www.kaggle.com/competitions/stars-classification/data}{Kaggle (competition)} &
1000 & 6  & 5  & 1  & target & 23.9 & Kaggle terms &
Natural Science — Hipparcos-style photometric and parallax features. \emph{Target = 1}: giant star; 0: dwarf star. \\
crimes\_arrest &
\href{https://www.kaggle.com/datasets/anacpricciardi/chicage-crimes-2025}{Kaggle (anacpricciardi)} &
1000 & 15 & 8  & 7  & target & 17.4 & Kaggle terms &
Crimes \& Justice — Chicago crimes 2025 incident records. \emph{Target = 1}: an arrest was made for this incident; 0: no arrest. \\
reading &
\href{https://data.mendeley.com/datasets/24wf5n2gk5/1}{Mendeley 24wf5n2gk5} &
582  & 16 & 0  & 16 & target & 61.7 & CC BY 4.0$^{\dagger}$ &
Education — undergraduate students' reading-habits survey. \emph{Target = 1}: student reads academic books most of the time; 0: reads non-academic books. \\
extrovert &
\href{https://www.kaggle.com/datasets/snigdha247/the-social-behavior-survey}{Kaggle (snigdha247)} &
1007 & 10 & 0  & 10 & target & 30.5 & Kaggle terms &
People \& Society — social-interaction preferences and emotional-response survey. \emph{Target = 1}: respondent maintains long-distance friendships through phone calls; 0: uses texting/social media or rarely keeps in touch. \\
\bottomrule
\end{tabular}%
}
\end{table*}

\subsection{Synthetic Datasets (MLP-prior)}
\label{app:synthetic_datasets}

The synthetic suite is generated by sampling random multi-layer perceptrons
of controlled depth and width, drawing inputs from $\mathcal{N}(0, I)$, and
labelling them by the sign of the network output. Seven difficulty levels
are obtained by varying layer count, hidden size, and input dimension; the
resulting datasets are used in RQ6 to study how performance scales with
task complexity. Table~\ref{tab:datasets_synthetic} gives the configuration
of each level, and Table~\ref{tab:intrinsic_dim} reports the empirically
estimated intrinsic dimensionality, averaged over seeds.

\begin{table*}[t]
\centering
\caption{MLP-prior difficulty levels.}
\label{tab:datasets_synthetic}
\begin{tabular}{@{}lccc@{}}
\toprule
\textbf{Difficulty} & \textbf{Layers} & \textbf{Features} & \textbf{Hidden size} \\
\midrule
MLP-0 & 0 & 5  & --- \\
MLP-1 & 1 & 10 & 10 \\
MLP-2 & 2 & 15 & 15 \\
MLP-3 & 3 & 20 & 20 \\
MLP-5 & 5 & 30 & 30 \\
MLP-7 & 7 & 40 & 40 \\
MLP-9 & 9 & 50 & 50 \\
\bottomrule
\end{tabular}
\end{table*}

\begin{table*}[t]
\centering
\caption{Empirical intrinsic dimensionality of the synthetic datasets, averaged across seeds.}
\label{tab:intrinsic_dim}
\begin{tabular}{@{}lr@{}}
\toprule
\textbf{Dataset} & \textbf{Avg.\ intrinsic dim.} \\
\midrule
mlp0 & 5.0241 \\
mlp1 & 9.3174 \\
mlp2 & 12.8327 \\
mlp3 & 16.0378 \\
mlp5 & 21.9469 \\
mlp7 & 27.4410 \\
mlp9 & 32.3677 \\
\bottomrule
\end{tabular}
\end{table*}

\subsection{LLM-synthetic Datasets}
\label{app:llm_synthetic_datasets}

For each of the 24 core real-world datasets we generate two LLM-synthetic
counterparts by prompting Qwen3-14B and GPT-4o-mini, respectively, to
fabricate tabular records matching the original schema and target
distribution. The resulting datasets are suffixed \texttt{\_ls\_qwen} and
\texttt{\_ls\_gpt} (e.g.\ \texttt{diabetes\_ls\_qwen},
\texttt{diabetes\_ls\_gpt}). They share the column structure and task
description of their real counterparts but contain no leaked instances
from the original data; this suite is the substrate for the
knowledge-source ablations in RQ2, RQ4, and RQ5.

\paragraph{Generator $\leftrightarrow$ backbone pairing.}
The \texttt{\_ls\_qwen} and \texttt{\_ls\_gpt} variants are not
interchangeable evaluation targets: each is paired one-to-one with
the model that generated it. The Qwen-generated variants are
evaluated only on the two Qwen ICL backbones (Qwen3-8B and
Qwen3-14B), and the GPT-generated variants only on GPT-4o-mini.
This pairing makes Prompt~3 a clean probe of each model's
\emph{own} internal task knowledge: the in-context examples,
decision rules, and inference backbone all come from the same
model family, so a non-trivial signal can be attributed to that
model's prior knowledge rather than to a distribution mismatch
between the generator and the evaluator. Cross-evaluation (e.g.\
running an \texttt{\_ls\_gpt} dataset on Qwen) is intentionally
omitted throughout the appendix.

\paragraph{Three-step generation procedure.}
For each \texttt{(real\_dataset, generator\_LLM)} pair the synthetic
dataset is produced through a chained, three-turn dialogue.\footnote{All
three turns share the same system message: \emph{``You are a statistical
expert who uses only clearly measurable data.''} This wording is meant
to nudge the generator away from heavily subjective survey-style
features.} The dialogue is structured so that each subsequent turn
receives the full text of the preceding answers as context, ensuring
that the schema introduced in turn~1 is honoured by the data-generating
function in turn~2 and by the labelling rules in turn~3.

\begin{itemize}
%\end{itemize}
  \item Turn 1 -- Schema design.
  
  The generator is asked to propose a
  \emph{prototype dataset} from the relevant high-level domain (e.g.\
  \emph{people and society}, \emph{healthcare}) such that the
  classification task matches that of the real counterpart. The model
  decides on the number of features, their semantics and admissible
  ranges/categories.
  \item Turn 2 -- Data-generating function.
  
  The generator is asked to
  return a Python function (using only \texttt{numpy} and
  \texttt{pandas}) that materialises rows according to the schema of
  turn~1. Numerical features are typically drawn from uniform or
  truncated normal distributions over the ranges declared in turn~1;
  categorical features are drawn uniformly over their declared support.
  \item Turn 3 -- Decision rules.
  
  The generator is asked for three
  measurable, threshold-based decision rules that map the features to a
  binary target label. These three rules form the \emph{ground-truth
  oracle} used to label rows produced in turn~2: a row is labelled
  positive if at least one of the three rules fires, and negative
  otherwise (the labelling logic visible inside the turn-3 code is what
  is actually executed).
\end{itemize}

\noindent The exact turn-1, turn-2 and turn-3 prompt templates, together
with one fully worked example (the \texttt{diabetes\_ls\_qwen}
dialogue), are reproduced in Section~\ref{app:llm_synth_prompts}.

\paragraph{Generation hyperparameters.}
Both generators are queried with their default decoding
configuration (temperature, top-$p$, and \texttt{max\_new\_tokens}
left at the provider's defaults; no custom sampling overrides).
We do not tune any decoding hyperparameters per dataset or per
turn. For each \texttt{(real\_dataset, generator\_LLM)} pair, we
retain all rows produced in turn~2 as the final synthetic dataset
(no rejection sampling, no rebalancing). The resulting class
imbalance is therefore the natural outcome of the turn-3 oracle
applied to the turn-2 distribution.

% \clearpage

% 
% =============================================================================
\section{Serialization formats}
\label{app:serialization_examples}

Each row is rendered into text using one of five serialization formats; the two
masked variants are used for the complexity grid.
\begin{itemize}
    \item \texttt{feat\_val}: A simple ``Feature = value'' listing. Example: ``Features are: age = 39, education = Bachelor, gain = 2174. Answer is 0.''
    \item \texttt{feat\_val\_mask}: A ``Feature = value'' listing with column names masked by proxy names. Example: ``Features are: x\_1 = 39, x\_2 = Bachelor, x\_3 = 2174. Answer is: y = 0.''
    \item \texttt{markdown}: A Markdown-formatted table.
    \item \texttt{markdown\_mask}: A Markdown table with feature (column-name) masking.
    \item \texttt{html}: An HTML-formatted table.
\end{itemize}

% \clearpage
\section{Prompts}
\label{app:prompts}
% =============================================================================

We evaluate every model under three prompt regimes, each isolating a
different combination of knowledge sources. All three regimes share a
common system message that instructs the model to emit a single token,
\texttt{0} or \texttt{1}, with no additional text. The variable component
across regimes is a per-dataset \emph{task description} (which embeds the
target semantics) and an optional \emph{decision-rule block} (which
embeds external expert knowledge synthesised by an auxiliary LLM).

Below we give the template for each regime followed by one fully
instantiated example. Full per-dataset task descriptions and decision
rules used in the experiments are released with the paper's code
repository.

%-------------------------------------------------------------------
\subsection{Prompt 1 — Contextual (full task description, no examples)}
\label{app:prompt1}

\textbf{Template.}\quad The contextual prompt describes the task,
explains what the input features represent, and defines the meaning of
both target labels. The features are then serialised one of five ways
(see Section~\ref{sec:3}).
\begin{quote}\small\itshape
For the given input features \textlangle\textit{of [entity]}\textrangle\
you need to predict \textlangle\textit{[task in natural language]}\textrangle.
Your output should be just a single number representing a binary class
prediction: 0 (\textlangle\textit{[negative-class meaning]}\textrangle)
or 1 (\textlangle\textit{[positive-class meaning]}\textrangle).
Do not predict any other tokens, only 0 or 1.
\end{quote}

\textbf{Example (\texttt{diabetes}).}
\begin{quote}\small
For the given input features you need to predict, based on diagnostic
measurements, whether a patient has diabetes. Your output should be
just a single number representing a binary class prediction:
0 (does not have diabetes) or 1 (has diabetes). Do not predict any
other tokens, only 0 or 1.
\end{quote}

%-------------------------------------------------------------------
\subsection{Prompt 2 — Context-free (query only)}
\label{app:prompt2}

\textbf{Template.}\quad The context-free prompt strips every domain hint
from the task. The model sees only the serialised features and a
generic two-class instruction; no entity, no target semantics, no
column meaning. This is the prompt used to isolate the contribution of
in-context demonstrations from prior knowledge.
\begin{quote}\small\itshape
For the given input features you need to classify rows into two
classes: 0 or 1. Do not predict any other tokens, only 0 or 1.
\end{quote}

\textbf{Example.}\quad Identical for every dataset — the verbatim
template above.

%-------------------------------------------------------------------
\subsection{Prompt 3 — Knowledge-augmented (task description + decision rules)}
\label{app:prompt3_expert_note}
This knowledge-augmented prompt defines the \emph{expert} regime referred to in the main text: it augments the task description with explicit, human-readable decision rules.
\label{app:prompt3}

\textbf{Template.}\quad The knowledge-augmented prompt extends
\textbf{Prompt~1} with an additional block listing the \emph{decision
rules by which the classes were generated}. These rules are produced by
an auxiliary LLM (Qwen3-8B for the \texttt{\_ls\_qwen} suite and
DeepSeek for the \texttt{\_ls\_gpt} suite) as a synthetic-data
annotation step; they are \emph{not} ground-truth rules for the real
datasets and serve as a structured prior of external expert knowledge.
\begin{quote}\small\itshape
\textlangle\textit{Prompt~1 task description}\textrangle\quad
Here are the decision rules by which the classes were generated:
\textlangle\textit{[rule block --- typically 3 rules expressed as
threshold conditions or short Python functions]}\textrangle
\end{quote}

\textbf{Example (\texttt{diabetes}, decision rules synthesised by Qwen3-8B).}
\begin{quote}\small
For the given input features you need to predict, based on diagnostic
measurements, whether a patient has diabetes. Your output should be just
a single number representing a binary class prediction: 0 (does not have
diabetes) or 1 (has diabetes). Do not predict any other tokens, only 0
or 1. Here are the decision rules by which the classes were generated:

\textbf{Rule 1 (High Glucose + High BMI $\rightarrow$ Diabetes Risk).}
If \texttt{Glucose > 120} AND \texttt{BMI > 30}, then
\texttt{Diabetes = 1}.

\textbf{Rule 2 (Age + Family History $\rightarrow$ Increased Risk).}
If \texttt{Age > 50} AND \texttt{FamilyHistory == 1}, then
\texttt{Diabetes = 1}.

\textbf{Rule 3 (High Insulin + Low Physical Activity $\rightarrow$
Diabetes Risk).} If \texttt{Insulin > 150} AND
\texttt{PhysicalActivity < 100}, then \texttt{Diabetes = 1}.
\end{quote}

\paragraph{Coverage of Prompt~3.}
Prompt~3 is applied \emph{only to the LLM-synthetic suite} (the
\texttt{\_ls\_qwen} and \texttt{\_ls\_gpt} variants of the 24 core
datasets), because the decision rules are valid only for data generated
under those same rules; applying them to the real datasets would
introduce label-rule mismatch. This explains why Prompt~3 columns are
empty for real and new-suite datasets throughout the per-domain tables
in Appendix~\ref{app:rq3}.

\paragraph{Per-dataset prompt text.}
Full prompt texts for all 24 core datasets, all 8 new-suite datasets,
and all decision-rule blocks (one per dataset $\times$ rule source) are
included verbatim in the code repository accompanying the paper, under
\texttt{prompts/prompt1.json}, \texttt{prompts/prompt2.json}, and
\texttt{prompts/prompt3\_\textlangle qwen\textbar gpt\textrangle.json}.

\subsection{Prompts used to generate the LLM-synthetic suite}
\label{app:llm_synth_prompts}

The LLM-synthetic suite described in Section~\ref{app:llm_synthetic_datasets} is
produced through a three-turn chained dialogue. All three turns share
the same system message:

\begin{quote}\itshape
You are a statistical expert who uses only clearly measurable data.
\end{quote}

\noindent The user messages are templated on the real dataset's domain
$D$ (e.g.\ \emph{healthcare}, \emph{people and society}) and the
real-task description $T$ (e.g.\ \emph{predict based on diagnostic
measurements whether a patient has diabetes}). The literal strings are
as follows.

\paragraph{Turn 1 — Schema design.}
\begin{quote}\small\itshape\noindent\rule{\linewidth}{0.4pt}\par\smallskip
Generate a prototype dataset from the $D$ domain. Use any number of
features, the classification task should be to $T$.
\par\smallskip\noindent\rule{\linewidth}{0.4pt}\end{quote}

\paragraph{Turn 2 — Data-generating function.}
The verbatim text of the answer to turn 1 (denoted $A_1$) is appended
to the user message before turn 2 is sent:
\begin{quote}\small\itshape\noindent\rule{\linewidth}{0.4pt}\par\smallskip
Generate python functions to create a dataset with the same features
following the given prototype.\\
$A_1$
\par\smallskip\noindent\rule{\linewidth}{0.4pt}\end{quote}

\paragraph{Turn 3 — Decision rules.}
The verbatim text of $A_2$ is appended to the user message before turn
3 is sent:
\begin{quote}\small\itshape\noindent\rule{\linewidth}{0.4pt}\par\smallskip
Generate python functions with 3 decision rules for determining a
target label based on the features of this $D$ dataset.\\
$A_2$
\par\smallskip\noindent\rule{\linewidth}{0.4pt}\end{quote}

\paragraph{Materialisation.} After turn 3 we extract the Python code
blocks from $A_2$ (data-generating function) and $A_3$ (three decision
rules), execute them in a sandboxed interpreter, and assemble the final
synthetic table. A row is labelled positive iff at least one of the
three turn-3 rules fires and negative otherwise.

\paragraph{Worked example (diabetes, Qwen3-14B).} The full transcripts
of $A_1$, $A_2$ and $A_3$ for the \texttt{diabetes\_ls\_qwen} dialogue
(used as the running illustration in Section~\ref{app:llm_synth_prompts}) are
reproduced verbatim in the supplementary material accompanying the
paper. Briefly: turn~1 proposed an 11-feature schema covering standard
diabetic-risk markers (\texttt{Age}, \texttt{BMI}, \texttt{Glucose},
\texttt{Insulin}, etc.); turn~2 returned a NumPy/Pandas function that
draws each feature from a clipped normal or uniform distribution;
turn~3 emitted three threshold rules — (i)~\texttt{Glucose~$>$~120 AND
BMI~$>$~30}, (ii)~\texttt{Age~$>$~50 AND FamilyHistory~==~1},
(iii)~\texttt{Insulin~$>$~150 AND PhysicalActivity~$<$~100} — which are
OR-combined into the final binary label.

% \clearpage

% =============================================================================
\section{Statistical Methodology}
\label{app:stats_methods}
% =============================================================================

\subsection{Wilcoxon signed-rank test against the random baseline}
\label{app:chance_test}
For binary classification, random ranking has expected ROC--AUC $0.5$.
Because the experiment stores aggregate metrics rather than per-example
prediction scores, we do not construct an empirical null distribution and do
not apply a test that requires individual predictions. Instead, the dataset is
the unit of analysis, following the standard multiple-dataset comparison
protocol of \citet{demsar2006statistical}.

For model $m$ and classic dataset $j$, let
$\overline{A}_{mj}$ be the zero-shot ROC--AUC averaged over the available
independent runs (at most five) in the primary full-context generation
configuration. We form the 24 paired differences
\begin{equation}
    d_{mj}=\overline{A}_{mj}-0.5
\end{equation}
and test
$H_0:\operatorname{median}_j(d_{mj})\leq 0$ against the one-sided alternative
$H_1:\operatorname{median}_j(d_{mj})>0$ with the Wilcoxon signed-rank test.
Zero differences are omitted from the signed-rank statistic in the standard
way.

This construction deliberately avoids treating the at most five repeated runs
as if they were many independent test examples. The independent units are the
benchmark datasets; averaging runs only reduces sensitivity to demonstration
sampling within each dataset. For each model we report the median paired
difference and the matched-pairs rank-biserial correlation. Since five models
are tested against the same null, raw $p$-values are adjusted with Holm's
step-down procedure~\cite{holm1979simple}, controlling the family-wise error
rate at $\alpha=0.05$.

\subsection{DeLong's test and Stouffer's method of $p$-value aggregation}
\label{app:del_test}
For each dataset and each specific experimental condition (defined by Model,
Mode, and a specific $k$ value), the statistical significance of the difference
between the $k$-shot ROC-AUC and the corresponding zero-shot ROC-AUC is
assessed. This is performed using \textbf{DeLong's test} for two correlated ROC
curves \cite{delong1988comparing}. This non-parametric test is specifically
designed to compare AUCs derived from the same set of test instances, making
it ideal for this paired, within-dataset comparison.
The $p$-values obtained from DeLong's test across all datasets for a given
experimental condition are not independent estimates but are observations of
the same underlying $k$-shot effect across different tasks. To obtain a global
assessment of significance for each condition, these $p$-values are aggregated
using Stouffer's method.

% \clearpage

% =============================================================================
\section{Visual results per research question}
\label{app:rq_figures}
% =============================================================================

This appendix collects the principal figures supporting the
research-question analyses of the main text, one block per RQ\,1--6.
Each block reproduces the headline plot(s) on which the corresponding
conclusions rely; the underlying numbers and statistical tests are
reported in the main text and in the per-domain tables of
Section~\ref{app:rq3}.

% -----------------------------------------------------------------------------
\subsection{RQ\,1 --- Zero-shot competitiveness against TabPFN-3 Plus}
\label{app:rq1}
% -----------------------------------------------------------------------------

RQ\,1 asks whether LLMs are competitive in zero-shot tabular
classification. Figure~\ref{fig:rq1_radar} (Appendix~\ref{app:key_figures})
reports the per-domain comparison against 16-shot TabPFN-3 Plus; the aggregate classic vs.\
post-cutoff results are reported in Table~\ref{tab:h1a_zero_shot_vs_tabpfn_all}.

\begin{table*}[!htbp]
\sisetup{separate-uncertainty=true,table-align-uncertainty=true,table-figures-uncertainty=1}
\centering
\scriptsize
\setlength{\tabcolsep}{2pt}
\resizebox{\textwidth}{!}{%
\begin{tabular}{lcccccc}
\toprule
Dataset Group & GPT-4o-mini (gen) & Qwen3-1.7B (gen) & Qwen3-8B (gen) & Qwen3-14B (gen) & GLM-4.7 Flash (gen) & TabPFN-3 Plus (16-shot) \\
\midrule
All & \num{0.722 \pm 0.008} & \num{0.605 \pm 0.007} & \num{0.684 \pm 0.001} & \num{0.693 \pm 0.000} & \num{0.649 \pm 0.028} & \num{0.732 \pm 0.062} \\
Classic & \num{0.731 \pm 0.009} & \num{0.608 \pm 0.007} & \num{0.689 \pm 0.000} & \num{0.699 \pm 0.000} & \num{0.667 \pm 0.030} & \num{0.736 \pm 0.060} \\
New & \num{0.664 \pm 0.008} & \num{0.587 \pm 0.005} & \num{0.657 \pm 0.001} & \num{0.657 \pm 0.000} & \num{0.607 \pm 0.023} & \num{0.693 \pm 0.073} \\
\bottomrule
\end{tabular}
}%
\caption{Zero-shot LLMs vs.\ TabPFN-3 Plus (16-shot), broken down by dataset group: \emph{All} (32 datasets), \emph{Classic} (24 established benchmarks), and \emph{New} (8 post-cutoff datasets).}
\label{tab:h1a_zero_shot_vs_tabpfn_all}
\end{table*}

\begin{figure*}[t]
    \centering
    \includegraphics[width=0.94\textwidth]{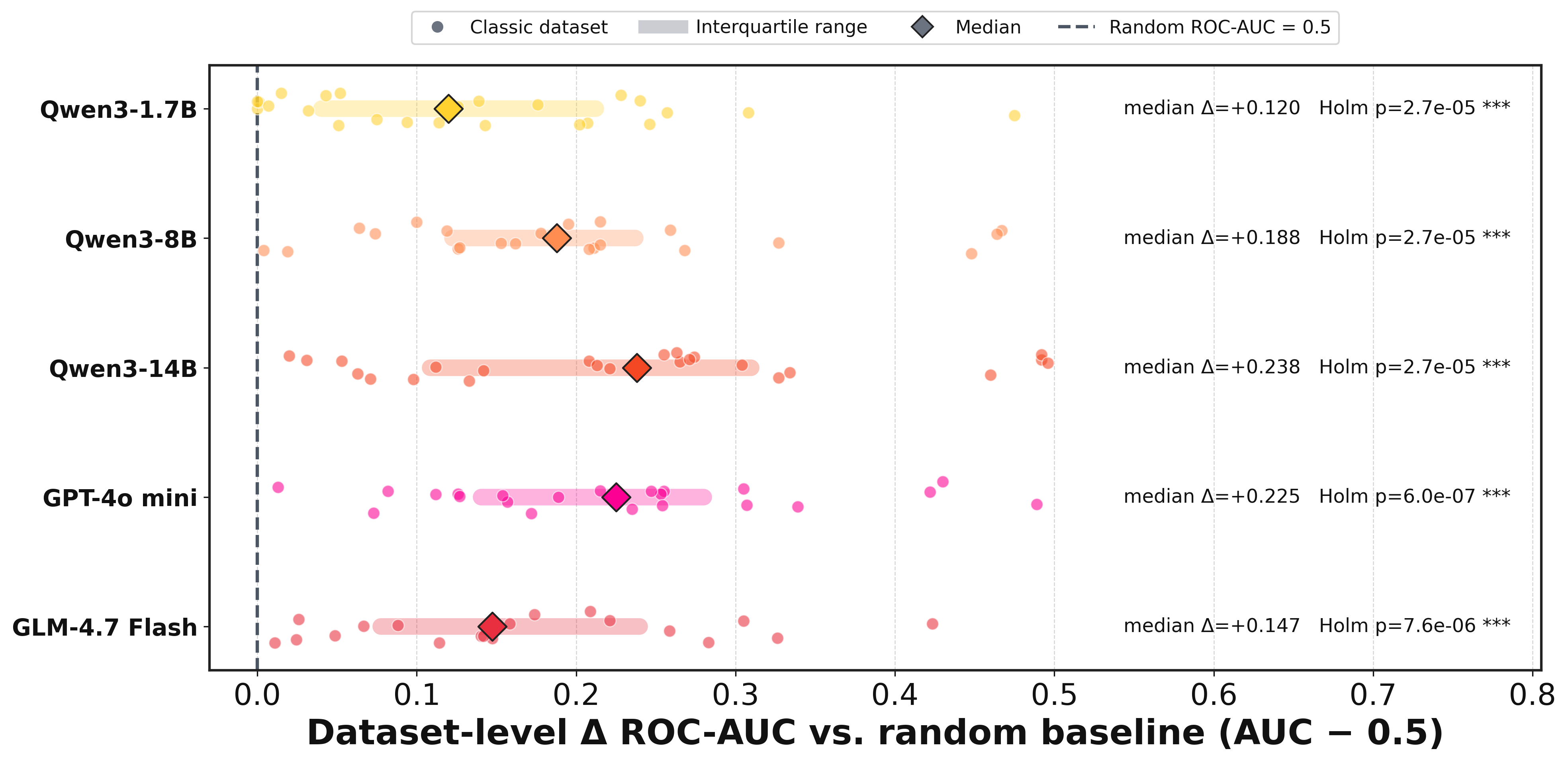}
    \caption{Zero-shot performance relative to random guessing on the 24
    classic datasets. Each small point is one dataset-level mean ROC--AUC
    difference from $0.5$; diamonds mark medians and thick segments show
    interquartile ranges. Reported $p$-values come from one-sided Wilcoxon
    signed-rank tests and are Holm-adjusted across the five models. GLM-4.7-Flash
    uses the available classic-dataset subset.}
    \label{fig:q1_zero_shot_vs_random}
    \Description{Dataset-level zero-shot ROC-AUC differences from chance for
    each model, with medians, interquartile ranges, and Holm-adjusted Wilcoxon
    significance values.}
\end{figure*}

\begin{table}[!htbp]
\centering
\footnotesize
\setlength{\tabcolsep}{4pt}
\caption{Stage-1 recognition outcomes per model (out of 24 datasets),
supporting the memorization-probe discussion in
Section~\ref{sec:q1_h1a}.}
\label{tab:recognition}
\begin{tabularx}{\linewidth}{@{}>{\raggedright\arraybackslash}X
  >{\centering\arraybackslash}p{1.4cm}
  >{\centering\arraybackslash}p{1.4cm}
  >{\centering\arraybackslash}p{1.4cm}@{}}
\toprule
Outcome & Qwen3-14B & GPT-4o-mini & GLM-4.7 Flash \\
\midrule
Leaked & 11 & 3 & 9 \\
Knows name only & 2 & 9 & 4 \\
Not recognized & 11 & 12 & 11 \\
\bottomrule
\end{tabularx}
\end{table}

\begin{table*}[!htbp]
\sisetup{
  separate-uncertainty = true,
  table-figures-decimal = 3,
}
\centering
\small
\setlength{\tabcolsep}{5pt}
\renewcommand{\arraystretch}{1.1}
 
\caption{Stage-2 cell-reconstruction results for datasets flagged at
the recognition stage. For each (model, dataset) pair, we report the
fraction of probed cells whose value the model reconstructs within the
correctness window (\emph{LLM}), the empirical random-baseline accuracy
computed from the column's marginal distribution under the same
window (\emph{Random}), and the resulting memorization \emph{lift}.
Verdict thresholds: $\text{lift} > 0.20$ -- \emph{leaked}; $> 0.10$ --
\emph{likely leaked}; $> 0.05$ -- \emph{suspicious}; otherwise -- \emph{no
leak signal}. Dashes indicate the dataset was not flagged at recognition
for that model and therefore not probed.}
\label{tab:reconstruction}
 
\small
\setlength{\tabcolsep}{3pt}
\begin{tabular}{@{}l ccc c ccc c >{\raggedright\arraybackslash}p{2.7cm}@{}}
\toprule
& \multicolumn{3}{c}{\textbf{Qwen3-14B}} && \multicolumn{3}{c}{\textbf{GPT-4o-mini}} && \\
\cmidrule(lr){2-4} \cmidrule(lr){6-8}
Dataset & LLM & Random & Lift && LLM & Random & Lift && Verdict \\
\midrule
adult                 & 0.420 & 0.385 & $+0.035$ && 0.530 & 0.385 & $\mathbf{+0.145}$ && \texttt{adult}: likely leaked (GPT) \\
biodegr               & 0.390 & 0.468 & $-0.078$ && --    & --    & --       && no leak signal \\
cancer                & 0.010 & 0.382 & $-0.372$ && --    & --    & --       && no leak signal \\
compas                & 0.420 & 0.597 & $-0.177$ && --    & --    & --       && no leak signal \\
credit                & 0.050 & 0.443 & $-0.393$ && --    & --    & --       && no leak signal \\
diabetes              & 0.160 & 0.160 & $+0.000$ && 0.090 & 0.160 & $-0.070$ && no leak signal \\
san\_francisco\_crimes  & 0.140 & 0.381 & $-0.241$ && --    & --    & --       && no leak signal \\
tae                   & 0.090 & 0.376 & $-0.286$ && --    & --    & --       && no leak signal \\
telco                 & 0.470 & 0.414 & $\mathit{+0.056}$ && --    & --    & --       && \texttt{telco}: suspicious (Qwen) \\
transfusion           & 0.070 & 0.235 & $-0.165$ && 0.040 & 0.235 & $-0.195$ && no leak signal \\
vote                  & 0.190 & 0.486 & $-0.296$ && --    & --    & --       && no leak signal \\
\bottomrule
\end{tabular}
\end{table*}

\paragraph{Dedicated LLM-tabular pipelines vs.\ direct prompting.}
Table~\ref{tab:gpt_methods_vs_shots} reports the full shot grid for direct
prompting against FeatLLM~\cite{han2024featllm} and LLM Trees~\cite{knauer2025llmtrees}
on GPT-4o-mini, supporting the robustness check reported in
Section~\ref{sec:q1_pipelines}: FeatLLM stays $0.02$--$0.03$ ROC--AUC lower
until nearly converging at 64 shots ($0.789$ vs.\ $0.793$), and LLM Trees
is substantially weaker ($0.594$) and shot-independent.

\begin{table*}[!t]
\centering
\small
\begingroup
\sisetup{
  separate-uncertainty = true,
  table-align-uncertainty = true,
  table-figures-uncertainty = 3,
}
\setlength{\tabcolsep}{8pt}

\resizebox{\linewidth}{!}{%
\begin{tabular}{lcccccc}
\toprule
Method
& 0-shot
& 4-shot
& 8-shot
& 16-shot
& 32-shot
& 64-shot \\
\midrule
Direct prompting
& \num{0.728 \pm 0.118}
& \num{0.752 \pm 0.143}
& \num{0.770 \pm 0.132}
& \num{0.774 \pm 0.135}
& \num{0.786 \pm 0.135}
& \num{0.793 \pm 0.134} \\

FeatLLM
& \num{0.703 \pm 0.110}
& \num{0.735 \pm 0.111}
& \num{0.745 \pm 0.126}
& \num{0.749 \pm 0.124}
& \num{0.766 \pm 0.120}
& \num{0.789 \pm 0.127} \\

LLM Trees
& \num{0.594 \pm 0.078}
& \num{0.594 \pm 0.078}
& \num{0.594 \pm 0.078}
& \num{0.594 \pm 0.078}
& \num{0.594 \pm 0.078}
& \num{0.594 \pm 0.078} \\
\bottomrule
\end{tabular}
}%

\caption{Dedicated LLM-tabular pipelines vs.\ direct prompting on the classic
suite with GPT-4o-mini, by number of in-context shots. \emph{Direct prompting}
is the same setup as Section~\ref{sec:q1_h1a}/\ref{sec:q2_h2b}. Cells are mean~$\pm$~std ROC--AUC
across datasets.}
\label{tab:gpt_methods_vs_shots}
\endgroup
\end{table*}

% -----------------------------------------------------------------------------
\subsection{In-context information vs.\ prior knowledge}
\label{app:rq2}
% -----------------------------------------------------------------------------

The questions disentangles the contributions of pretrained prior knowledge and
in-context information by progressively removing contextual cues from
the prompt. Table~\ref{tab:h2b_context_configs} gives the full per-model
breakdown underlying the H2a discussion in Section~\ref{sec:q2_h2a};
Figure~\ref{fig:rq2_pk0shot} presents the same comparison in the main text.

\begin{table*}[ht]
\sisetup{
  separate-uncertainty = true,
  table-align-uncertainty = true,
  table-figures-uncertainty = 1,
}
\centering
\small
\setlength{\tabcolsep}{8pt}
\begin{tabular}{lccccc}
\toprule
Configuration & GPT-4o-mini & Qwen3-1.7B & Qwen3-8B & Qwen3-14B & GLM-4.7 Flash \\
\midrule
NoCols-NoContext & \num{0.625 \pm 0.013} & \num{0.606 \pm 0.004} & \num{0.579 \pm 0.006} & \num{0.596 \pm 0.002} & \num{0.564 \pm 0.025} \\
Cols-NoContext & \num{0.682 \pm 0.013} & \num{0.653 \pm 0.003} & \num{0.637 \pm 0.008} & \num{0.678 \pm 0.000} & \num{0.637 \pm 0.026} \\
NoCols-Context & \num{0.641 \pm 0.013} & \num{0.608 \pm 0.010} & \num{0.624 \pm 0.001} & \num{0.660 \pm 0.000} & \num{0.589 \pm 0.031} \\
Cols-Context & \num{0.728 \pm 0.011} & \num{0.640 \pm 0.007} & \num{0.711 \pm 0.000} & \num{0.733 \pm 0.000} & \num{0.677 \pm 0.032} \\
\bottomrule
\end{tabular}
\caption{Context configurations comparison across models (0-shot). \emph{Cols} indicates the presence of column names (schema indicators); \emph{Context} indicates the presence of the natural-language task description.}
\label{tab:h2b_context_configs}
\end{table*}

% -----------------------------------------------------------------------------
\subsection{Few-shot effectiveness}
\label{app:rq3}
% -----------------------------------------------------------------------------

The question quantifies the gains from few-shot demonstrations across
$k \in \{0, 4, 8, 16, 32, 64\}$. Table~\ref{tab:h2a_shot_scaling_classic}
gives the full per-model grid underlying the H2b discussion in
Section~\ref{sec:q2_h2b}; Figure~\ref{fig:rq3_shots_curve_full} presents
the corresponding trajectories in the main text.

\begin{table*}[!htbp]
\sisetup{
  separate-uncertainty = true,
  table-align-uncertainty = true,
  table-figures-uncertainty = 1,
}
\centering
\footnotesize
\setlength{\tabcolsep}{5pt}
\begin{tabular}{lcccccc}
\toprule
Model & 0-shot & 4-shot & 8-shot & 16-shot & 32-shot & 64-shot \\
\midrule
Qwen3-1.7B & \num{0.635 \pm 0.117} & \num{0.669 \pm 0.122} & \num{0.666 \pm 0.114} & \num{0.684 \pm 0.116} & \num{0.696 \pm 0.120} & \num{0.708 \pm 0.113} \\
Qwen3-8B & \num{0.700 \pm 0.126} & \num{0.735 \pm 0.138} & \num{0.756 \pm 0.128} & \num{0.769 \pm 0.130} & \num{0.780 \pm 0.126} & \num{0.790 \pm 0.127} \\
Qwen3-14B & \num{0.733 \pm 0.146} & \num{0.754 \pm 0.134} & \num{0.774 \pm 0.134} & \num{0.784 \pm 0.131} & \num{0.792 \pm 0.132} & \num{0.801 \pm 0.130} \\
GPT-4o mini & \num{0.725 \pm 0.117} & \num{0.763 \pm 0.132} & \num{0.774 \pm 0.131} & \num{0.779 \pm 0.133} & \num{0.786 \pm 0.135} & \num{0.792 \pm 0.134} \\
GLM-4.7 Flash & \num{0.667 \pm 0.111} & \num{0.722 \pm 0.130} & \num{0.736 \pm 0.137} & \num{0.748 \pm 0.133} & \num{0.741 \pm 0.117} & \num{0.723 \pm 0.160} \\
TabPFN & \num{0.500 \pm 0.000} & \num{0.696 \pm 0.137} & \num{0.721 \pm 0.137} & \num{0.772 \pm 0.142} & \num{0.814 \pm 0.136} & \num{0.845 \pm 0.121} \\
\bottomrule
\end{tabular}
\caption{Mean ROC-AUC $\pm$ standard deviation across the 24 classic datasets as the number of in-context examples increases. TabPFN has no 0-shot inference, so its 0-shot entry is fixed to random ROC-AUC $0.5$.}
\label{tab:h2a_shot_scaling_classic}
\end{table*}

% -----------------------------------------------------------------------------
\subsection{LLM-specific tabular baselines on the new-dataset bundle}
\label{app:new_datasets_llm_baselines}
% -----------------------------------------------------------------------------

To situate our direct in-context approach against other ways of using
LLMs for tabular prediction, we additionally compare it with two
LLM-specific tabular baselines, \textbf{FeatLLM} and \textbf{LLM-tree},
on the eight post-cutoff (\emph{new}) datasets introduced in
RQ\,1. Unlike our setup --- which classifies each test row directly from
the serialized prompt --- these methods use the LLM as an intermediate
step: FeatLLM prompts the LLM to extract discriminative decision
rules / binary features that a lightweight downstream predictor then
aggregates, while LLM-tree elicits a decision-tree-style classifier
guided by the LLM. Both baselines use GPT-4o-mini as the backbone. We
deliberately restrict this comparison to the \emph{new} datasets,
because they are released after the models' knowledge cutoff and thus
provide the cleanest, contamination-free setting in which to contrast
LLM-based methods.

Table~\ref{tab:new_datasets_llm_baselines_0shot} reports zero-shot
ROC--AUC per dataset. The direct in-context GPT-4o-mini classifier is
competitive with both LLM-specific baselines, leading on several
datasets (e.g.\ \texttt{stars}, \texttt{reading}, \texttt{crimes\_arrest})
while trailing on others (e.g.\ \texttt{callcenter}, \texttt{postpartum}),
and the open-weight Qwen3 models stay in the same range.
Table~\ref{tab:new_datasets_llm_baselines_fewshot} reports the few-shot
trajectory averaged over the new datasets; LLM-tree is omitted there
because it does not support few-shot prompting. FeatLLM is competitive
at low shot counts and is matched or modestly exceeded by the
mid-to-large in-context LLMs as the number of shots grows --- the same
few-shot trend observed on the classic suite (RQ\,3) therefore also
holds against an LLM-specific baseline on contamination-clean data.

\begin{table*}[t!]
\sisetup{
  separate-uncertainty = true,
  table-align-uncertainty = true,
  table-figures-uncertainty = 1,
}
\centering
\small
\setlength{\tabcolsep}{4pt}
\resizebox{\linewidth}{!}{%
\begin{tabular}{lcccccc}
\toprule
Dataset & GPT-4o-mini & Qwen3-1.7B & Qwen3-8B & Qwen3-14B & FeatLLM & LLM-tree \\
\midrule
stars & \num{0.995 \pm 0.001} & \num{0.616 \pm 0.001} & \num{0.982 \pm 0.000} & \num{0.990 \pm 0.000} & \num{0.959 \pm 0.024} & \num{0.631 \pm 0.120} \\
machine & \num{0.545 \pm 0.016} & \num{0.533 \pm 0.003} & \num{0.536 \pm 0.000} & \num{0.512 \pm 0.000} & \num{0.551 \pm 0.009} & \num{0.504 \pm 0.002} \\
callcenter & \num{0.522 \pm 0.010} & \num{0.586 \pm 0.001} & \num{0.510 \pm 0.000} & \num{0.543 \pm 0.000} & \num{0.638 \pm 0.052} & \num{0.557 \pm 0.011} \\
bank\_credit\_scoring & \num{0.707 \pm 0.004} & \num{0.527 \pm 0.001} & \num{0.529 \pm 0.001} & \num{0.712 \pm 0.000} & \num{0.697 \pm 0.028} & \num{0.579 \pm 0.083} \\
extrovert & \num{0.568 \pm 0.001} & \num{0.604 \pm 0.000} & \num{0.614 \pm 0.000} & \num{0.627 \pm 0.000} & \num{0.575 \pm 0.008} & \num{0.548 \pm 0.032} \\
reading & \num{0.797 \pm 0.003} & \num{0.500 \pm 0.000} & \num{0.745 \pm 0.002} & \num{0.800 \pm 0.000} & \num{0.620 \pm 0.029} & \num{0.746 \pm 0.075} \\
postpartum & \num{0.527 \pm 0.009} & \num{0.696 \pm 0.002} & \num{0.753 \pm 0.001} & \num{0.561 \pm 0.000} & \num{0.731 \pm 0.014} & \num{0.550 \pm 0.034} \\
crimes\_arrest & \num{0.650 \pm 0.005} & \num{0.630 \pm 0.013} & \num{0.590 \pm 0.000} & \num{0.514 \pm 0.000} & \num{0.619 \pm 0.079} & \num{0.531 \pm 0.011} \\
\bottomrule
\end{tabular}
}%
\caption{Zero-shot ROC--AUC on the eight post-cutoff \emph{new} datasets:
our direct in-context LLM classifiers vs.\ the LLM-specific tabular
baselines FeatLLM and LLM-tree (both run with GPT-4o-mini as backbone).
Entries are mean $\pm$ standard deviation.}
\label{tab:new_datasets_llm_baselines_0shot}
\end{table*}

\begin{table*}[t!]
\sisetup{
  separate-uncertainty = true,
  table-align-uncertainty = true,
  table-figures-uncertainty = 1,
}
\centering
\small
\setlength{\tabcolsep}{8pt}
\resizebox{\linewidth}{!}{%
\begin{tabular}{lcccccc}
\toprule
Model & 0-shot & 4-shot & 8-shot & 16-shot & 32-shot & 64-shot \\
\midrule
GPT-4o-mini & \num{0.664 \pm 0.008} & \num{0.711 \pm 0.035} & \num{0.720 \pm 0.035} & \num{0.723 \pm 0.039} & \num{0.723 \pm 0.040} & \num{0.730 \pm 0.037} \\
Qwen3-1.7B & \num{0.587 \pm 0.005} & \num{0.602 \pm 0.057} & \num{0.599 \pm 0.054} & \num{0.630 \pm 0.040} & \num{0.637 \pm 0.067} & \num{0.660 \pm 0.038} \\
Qwen3-8B & \num{0.657 \pm 0.001} & \num{0.695 \pm 0.032} & \num{0.705 \pm 0.038} & \num{0.716 \pm 0.035} & \num{0.723 \pm 0.033} & \num{0.739 \pm 0.027} \\
Qwen3-14B & \num{0.657 \pm 0.000} & \num{0.707 \pm 0.044} & \num{0.717 \pm 0.044} & \num{0.731 \pm 0.029} & \num{0.732 \pm 0.031} & \num{0.752 \pm 0.025} \\
FeatLLM & \num{0.674 \pm 0.038} & \num{0.668 \pm 0.051} & \num{0.680 \pm 0.054} & \num{0.701 \pm 0.057} & \num{0.703 \pm 0.054} & \num{0.727 \pm 0.049} \\
\bottomrule
\end{tabular}
}%
\caption{Few-shot ROC--AUC averaged over the \emph{new} datasets vs.\
shot count, including the FeatLLM baseline (GPT-4o-mini backbone).
LLM-tree is omitted because it does not support few-shot prompting.}
\label{tab:new_datasets_llm_baselines_fewshot}
\end{table*}

% -----------------------------------------------------------------------------
\subsection{Context contribution across shot regimes}
\label{app:rq4}
% -----------------------------------------------------------------------------

This questions extends Q1 to the few-shot regime: as the shot count $k$
grows, how much of the headline performance is still attributable to
the contextual cues in the prompt? Table~\ref{tab:delta_context_vs_nocontext}
gives the full per-shot $\Delta$ = \emph{Cols-Context} $-$
\emph{NoCols-NoContext} underlying the H2c discussion in
Section~\ref{sec:q2_h2c}.

\begin{table*}[ht]
\sisetup{
  separate-uncertainty = true,
  table-align-uncertainty = true,
  table-figures-uncertainty = 1,
}
\centering
\small
\setlength{\tabcolsep}{8pt}
\resizebox{\linewidth}{!}{%
\begin{tabular}{lcccccc}
\toprule
Model & 0-shot & 4-shot & 8-shot & 16-shot & 32-shot & 64-shot \\
\midrule
GPT-4o-mini & \num{+0.103 \pm 0.017} & \num{+0.061 \pm 0.065} & \num{+0.062 \pm 0.071} & \num{+0.047 \pm 0.067} & \num{+0.041 \pm 0.050} & \num{+0.032 \pm 0.038} \\
Qwen3-1.7B  & \num{+0.034 \pm 0.008} & \num{+0.019 \pm 0.080} & \num{+0.011 \pm 0.100} & \num{+0.026 \pm 0.094} & \num{+0.005 \pm 0.080} & \num{+0.001 \pm 0.077} \\
Qwen3-8B    & \num{+0.132 \pm 0.006} & \num{+0.062 \pm 0.075} & \num{+0.062 \pm 0.074} & \num{+0.043 \pm 0.070} & \num{+0.040 \pm 0.051} & \num{+0.023 \pm 0.038} \\
Qwen3-14B   & \num{+0.137 \pm 0.002} & \num{+0.060 \pm 0.071} & \num{+0.061 \pm 0.069} & \num{+0.036 \pm 0.064} & \num{+0.033 \pm 0.045} & \num{+0.026 \pm 0.036} \\
GLM-4.7 Flash & \num{+0.081 \pm 0.044} & \num{+0.018 \pm 0.054} & \num{+0.018 \pm 0.068} & \num{+0.020 \pm 0.054} & \num{+0.011 \pm 0.055} & \num{+0.007 \pm 0.052} \\
\bottomrule
\end{tabular}
}%
\caption{$\Delta$ ROC--AUC = \emph{Cols-Context} $-$
\emph{NoCols-NoContext} across shot counts on the classic suite.}
\label{tab:delta_context_vs_nocontext}
\end{table*}

Figure~\ref{fig:rq2_pk0shot} in the main text shows the four-way
context comparison of RQ\,2 at $k = 0$ and $k = 64$; the full
per-shot values underlying it, including $k \in \{4, 8, 16, 32\}$,
are given in Table~\ref{tab:delta_context_vs_nocontext} above. Together
they show that the gap
between full-context and minimal-context configurations shrinks
substantially with $k$, indicating partial substitution between
in-context information and few-shot demonstrations.

% -----------------------------------------------------------------------------
\subsection{Combining ICL entities on LLM-synthetic data}
\label{app:rq5}
% -----------------------------------------------------------------------------

\begin{figure}[ht!]
    \centering
    \includegraphics[width=\linewidth]{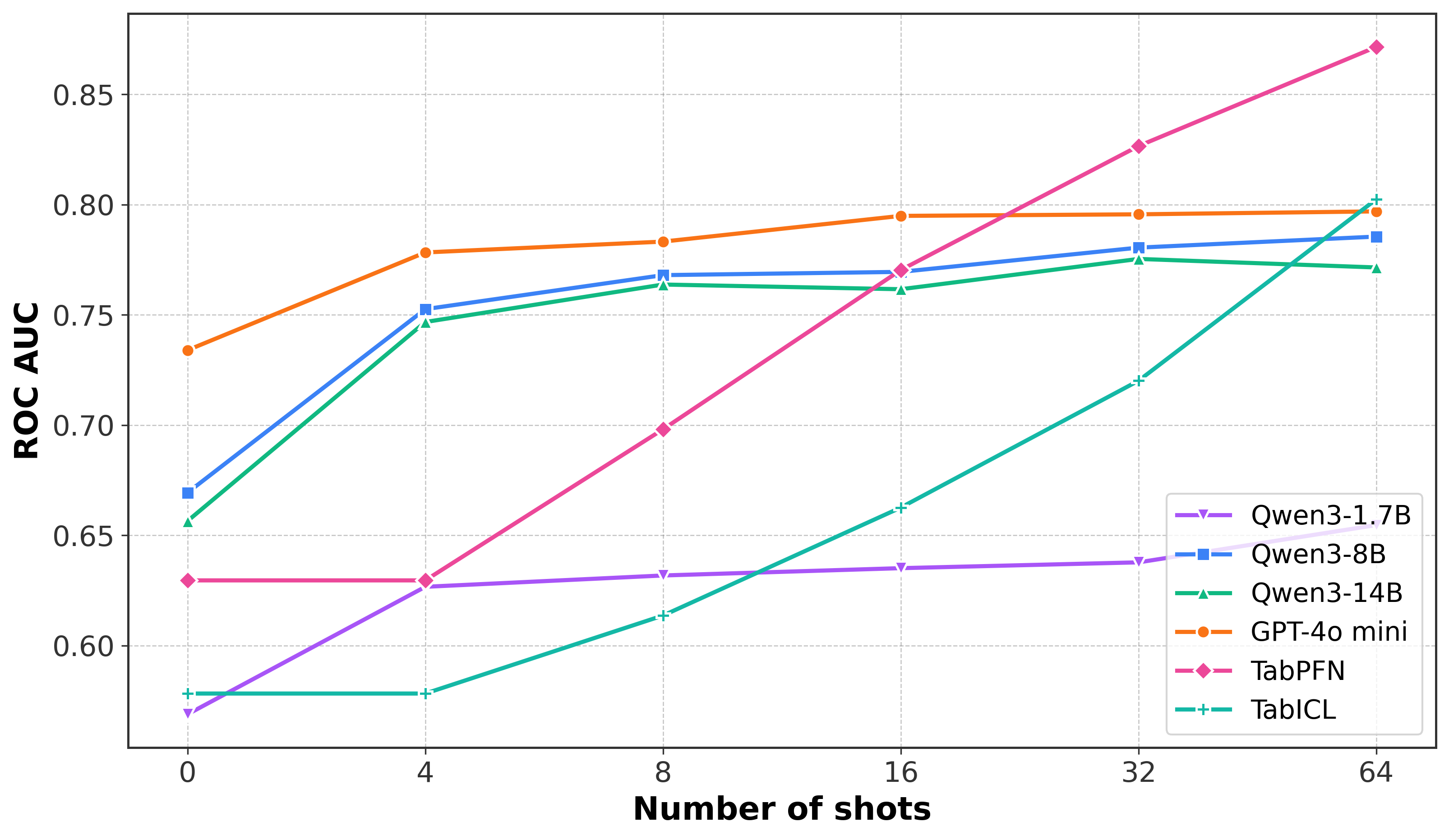}
    \caption{ROC--AUC vs.\ number of in-context shots on the
    LLM-synthetic suite. Performance saturates earlier than on real
    data because the model already aligns with the data-generating
    distribution at $k = 0$.}
    \label{fig:rq5_shots_curve}
    \Description{ROC--AUC vs.}
\end{figure}

\begin{figure}[ht!]
    \centering
    \includegraphics[width=\linewidth]{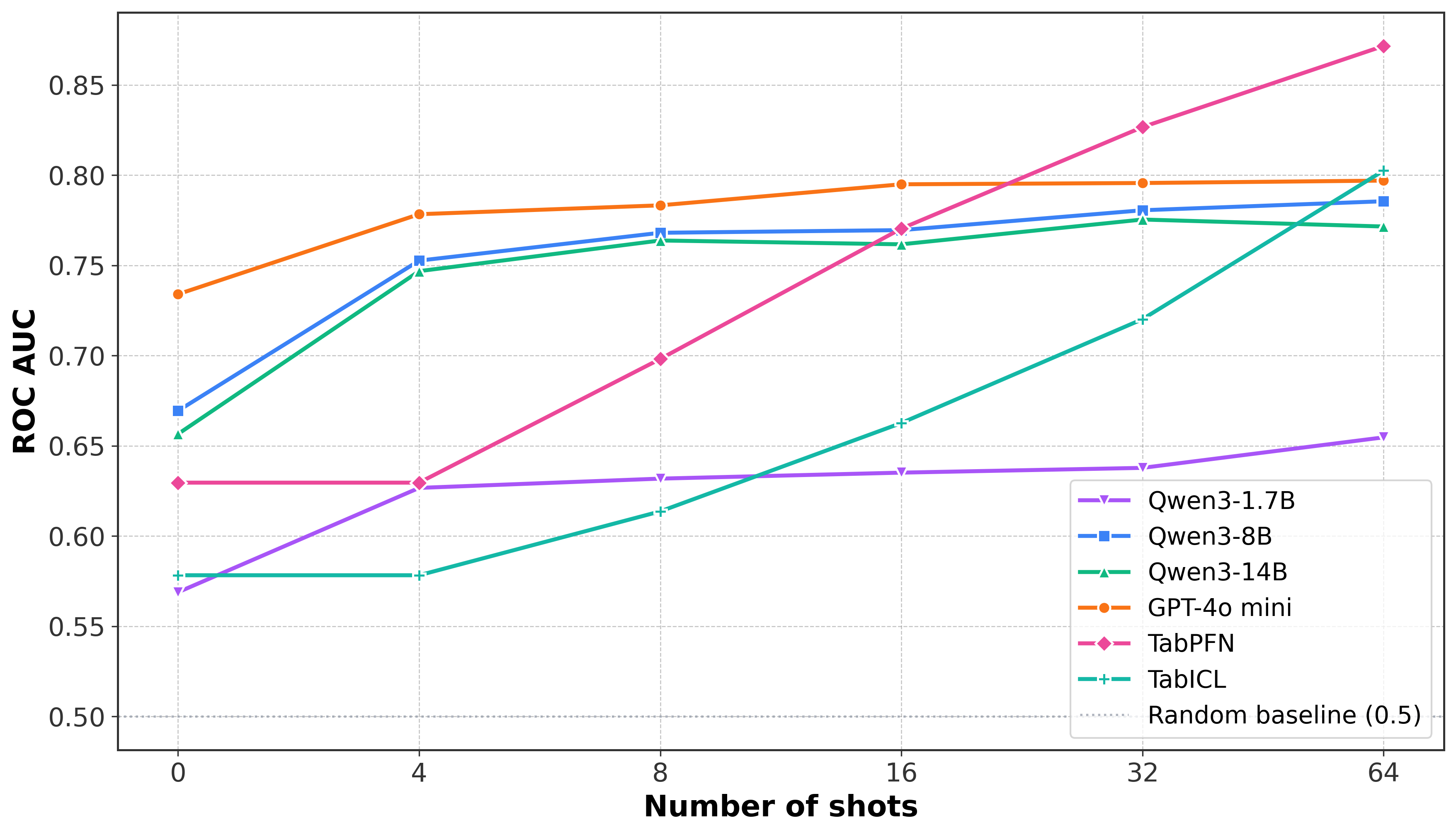}
    \caption{Cross-model comparison on the LLM-synthetic suite,
    including the foundation baselines TabPFN-3 Plus and TabICL. The relative
    ranking on LLM-synthetic differs from the ranking on real
    datasets, reflecting alignment between the generator and its own
    test distribution.}
    \label{fig:rq5_llms_comparison}
    \Description{Cross-model comparison on the LLM-synthetic suite, including the foundation baselines TabPFN-3 Plus and TabICL.}
\end{figure}

\begin{figure}[ht!]
    \centering
    \includegraphics[width=\linewidth]{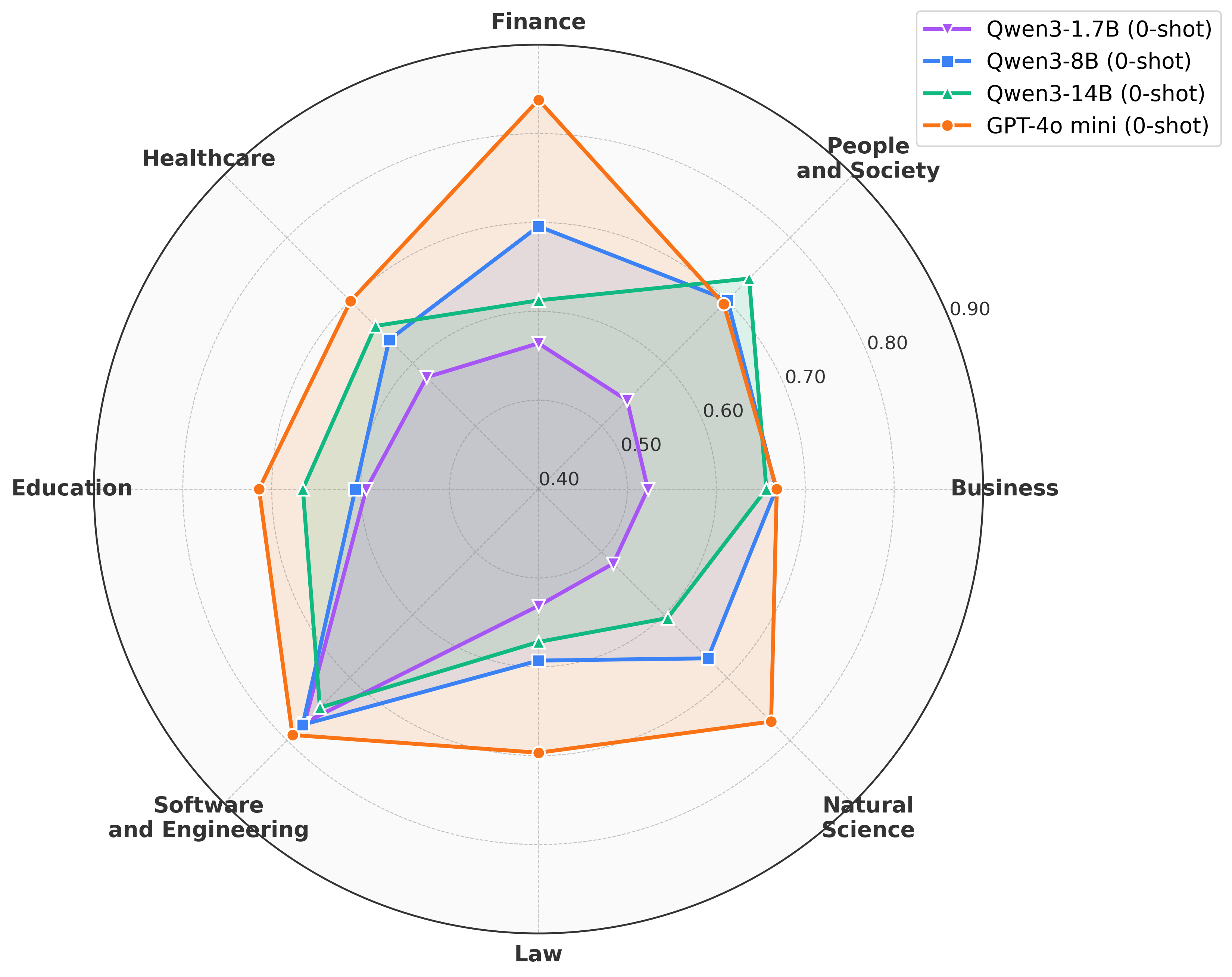}
    \caption{0-shot per-domain radar on the LLM-synthetic suite. All
    LLMs reach or exceed the foundation-model trajectories on the
    synthetic counterparts of their own training distribution.}
    \label{fig:rq5_radar_0}
    \Description{0-shot per-domain radar on the LLM-synthetic suite.}
\end{figure}

\begin{figure}[ht!]
    \centering
    \includegraphics[width=\linewidth]{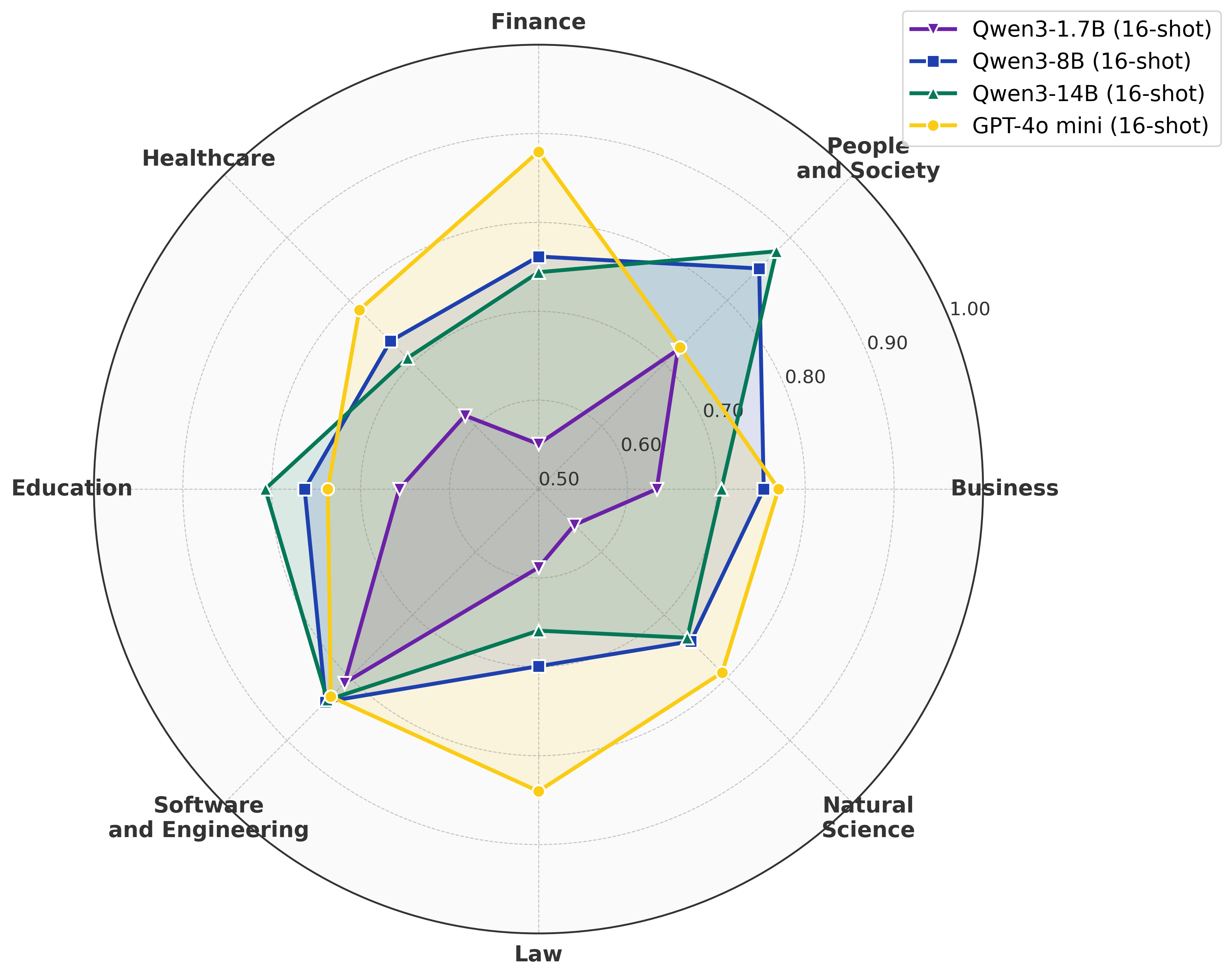}
    \caption{16-shot per-domain radar on the LLM-synthetic suite. The
    addition of demonstrations narrows the spread between models;
    smaller LLMs benefit the most relative to their 0-shot value.}
    \label{fig:rq5_radar_16}
    \Description{16-shot per-domain radar on the LLM-synthetic suite.}
\end{figure}

\begin{figure}[ht!]
    \centering
    \includegraphics[width=\linewidth]{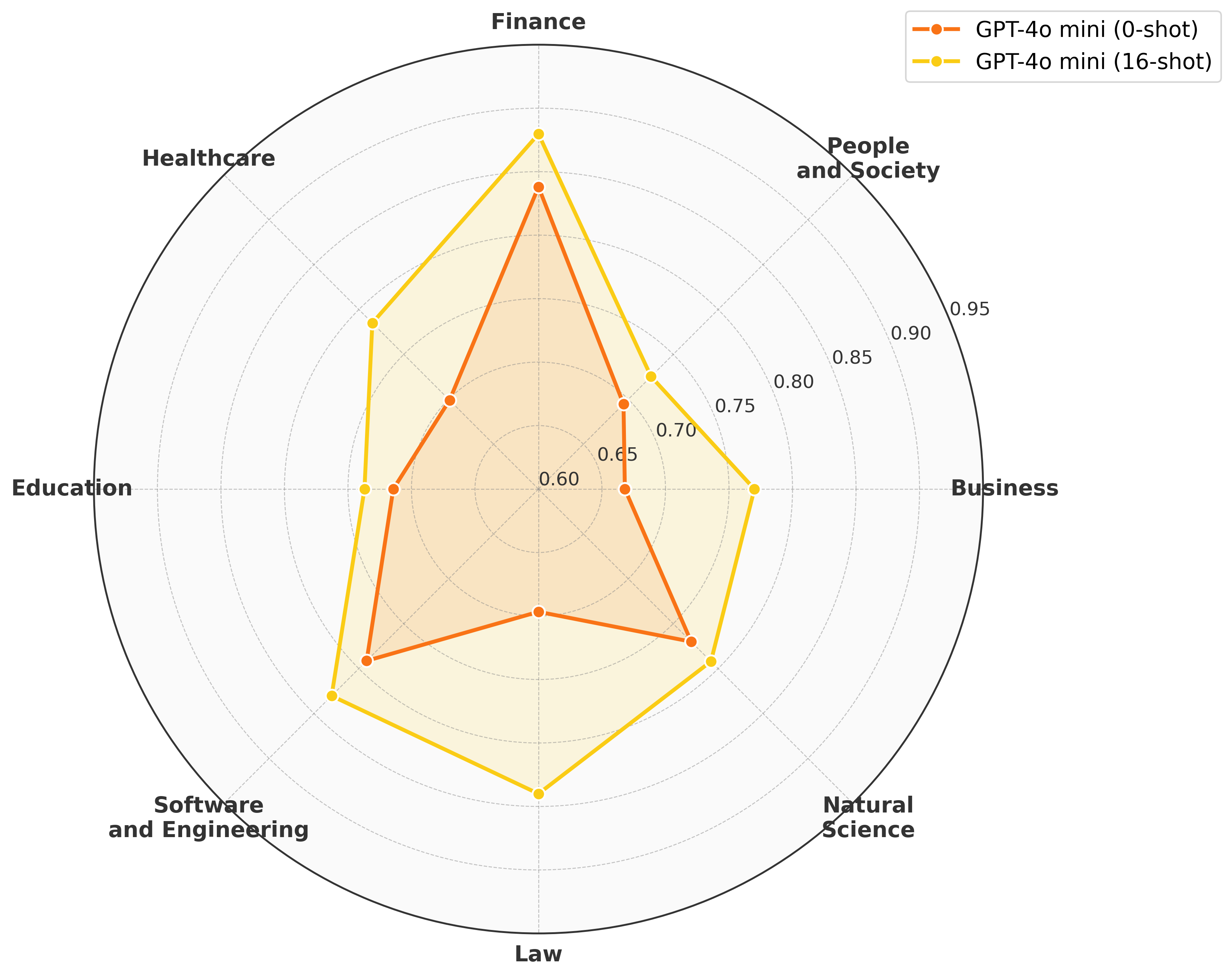}
    \caption{GPT-4o-mini: 0-shot vs.\ 16-shot per-domain radar on the
    LLM-synthetic suite. Confirms the same saturation pattern observed
    on real data, with a smaller absolute gain because the 0-shot
    baseline is already close to the model's ceiling on its own
    synthetic distribution.}
    \label{fig:rq5_gpt_radar}
    \Description{GPT-4o-mini: 0-shot vs.}
\end{figure}

Q3 asks whether combining in-context information, few-shot
demonstrations, and external (LLM-generated) decision rules improves
performance beyond using each source in isolation. We evaluate the
combinations on the \emph{LLM-synthetic} suite, where the model has
maximal alignment with the dataset's data-generating
process. Table~\ref{tab:h3_icl_combinations} gives the full comparison
underlying the H3 discussion in Section~\ref{sec:q3_h3}, and
Figures~\ref{fig:rq5_shots_curve}--\ref{fig:rq5_gpt_radar}
report shot-count curves, model comparison, and per-domain
radar charts at 0 and 16 shots.

% -----------------------------------------------------------------------------
\subsection{Dataset complexity and the few-shot ceiling}
\label{app:rq6}

\begin{table*}[!htbp]
\sisetup{
  separate-uncertainty = true,
  table-align-uncertainty = true,
  table-figures-uncertainty = 1,
}
\centering
\footnotesize
\setlength{\tabcolsep}{7pt}
\begin{tabular}{lcccc}
\toprule
Synthetic Dataset & Intrinsic Dim. & Qwen3-14B (0-shot) & GPT-4o-mini (0-shot) & GLM-4.7 Flash (0-shot) \\
\midrule
MLP0 & \num{4.830} & \num{0.602 \pm 0.050} & \num{0.608 \pm 0.083} & \num{0.536 \pm 0.016} \\
MLP1 & \num{9.012} & \num{0.561 \pm 0.042} & \num{0.596 \pm 0.063} & \num{0.543 \pm 0.025} \\
MLP2 & \num{12.480} & \num{0.545 \pm 0.022} & \num{0.563 \pm 0.020} & \num{0.520 \pm 0.009} \\
MLP3 & \num{15.705} & \num{0.578 \pm 0.043} & \num{0.603 \pm 0.064} & \num{0.540 \pm 0.017} \\
MLP5 & \num{21.544} & \num{0.534 \pm 0.046} & \num{0.546 \pm 0.033} & \num{0.531 \pm 0.014} \\
MLP7 & \num{27.023} & \num{0.541 \pm 0.019} & \num{0.533 \pm 0.014} & \num{0.520 \pm 0.010} \\
MLP9 & \num{31.994} & \num{0.516 \pm 0.008} & \num{0.526 \pm 0.013} & \num{0.525 \pm 0.003} \\
\bottomrule
\end{tabular}
\caption{Intrinsic dimensionality and zero-shot ROC-AUC on the synthetic MLP suite. Intrinsic dimensionality and model quality are averaged across five seeds per complexity level; uncertainty reports the standard deviation across seeds.}
\label{tab:h1b_mlp_complexity}
\end{table*}

\begin{table*}[ht]
\sisetup{
  separate-uncertainty = true,
  table-align-uncertainty = true,
  table-figures-uncertainty = 1,
}
\centering
\tiny
\setlength{\tabcolsep}{4pt}
\resizebox{\linewidth}{!}{%
\begin{tabular}{lcccccccccccc}
\toprule
Dataset & \multicolumn{3}{c}{GPT-4o-mini} & \multicolumn{3}{c}{Qwen3-8B} & \multicolumn{3}{c}{Qwen3-14B} & \multicolumn{3}{c}{GLM-4.7 Flash} \\
\cmidrule(lr){2-4} \cmidrule(lr){5-7} \cmidrule(lr){8-10} \cmidrule(lr){11-13}
 & 0-shot & 4-shot & 16-shot & 0-shot & 4-shot & 16-shot & 0-shot & 4-shot & 16-shot & 0-shot & 4-shot & 16-shot \\
\midrule
MLP0 & \num{0.608 \pm 0.004} & \num{0.645 \pm 0.065} & \num{0.708 \pm 0.055} & \num{0.564 \pm 0.001} & \num{0.639 \pm 0.070} & \num{0.713 \pm 0.059} & \num{0.602 \pm 0.000} & \num{0.688 \pm 0.088} & \num{0.748 \pm 0.059} & \num{0.536 \pm 0.023} & \num{0.605 \pm 0.057} & \num{0.685 \pm 0.057} \\
MLP1 & \num{0.596 \pm 0.007} & \num{0.593 \pm 0.023} & \num{0.578 \pm 0.033} & \num{0.562 \pm 0.004} & \num{0.587 \pm 0.032} & \num{0.568 \pm 0.051} & \num{0.562 \pm 0.000} & \num{0.584 \pm 0.032} & \num{0.579 \pm 0.053} & \num{0.543 \pm 0.014} & \num{0.555 \pm 0.031} & \num{0.563 \pm 0.039} \\
MLP2 & \num{0.563 \pm 0.010} & \num{0.558 \pm 0.023} & \num{0.566 \pm 0.020} & \num{0.547 \pm 0.004} & \num{0.564 \pm 0.022} & \num{0.580 \pm 0.030} & \num{0.535 \pm 0.000} & \num{0.564 \pm 0.029} & \num{0.577 \pm 0.028} & \num{0.520 \pm 0.011} & \num{0.552 \pm 0.026} & \num{0.550 \pm 0.025} \\
MLP3 & \num{0.603 \pm 0.008} & \num{0.569 \pm 0.035} & \num{0.581 \pm 0.026} & \num{0.566 \pm 0.011} & \num{0.576 \pm 0.027} & \num{0.573 \pm 0.027} & \num{0.583 \pm 0.001} & \num{0.587 \pm 0.041} & \num{0.587 \pm 0.031} & \num{0.540 \pm 0.019} & \num{0.558 \pm 0.026} & \num{0.574 \pm 0.030} \\
MLP5 & \num{0.546 \pm 0.006} & \num{0.542 \pm 0.014} & \num{0.541 \pm 0.013} & \num{0.527 \pm 0.007} & \num{0.536 \pm 0.016} & \num{0.538 \pm 0.018} & \num{0.541 \pm 0.000} & \num{0.537 \pm 0.015} & \num{0.543 \pm 0.026} & \num{0.531 \pm 0.012} & \num{0.537 \pm 0.023} & \num{0.536 \pm 0.022} \\
MLP7 & \num{0.533 \pm 0.005} & \num{0.538 \pm 0.019} & \num{0.541 \pm 0.012} & \num{0.523 \pm 0.006} & \num{0.534 \pm 0.018} & \num{0.535 \pm 0.018} & \num{0.537 \pm 0.000} & \num{0.534 \pm 0.017} & \num{0.540 \pm 0.019} & \num{0.520 \pm 0.012} & \num{0.524 \pm 0.016} & \num{0.510 \pm 0.011} \\
MLP9 & \num{0.526 \pm 0.007} & \num{0.526 \pm 0.012} & \num{0.525 \pm 0.009} & \num{0.526 \pm 0.009} & \num{0.525 \pm 0.016} & \num{0.529 \pm 0.014} & \num{0.521 \pm 0.000} & \num{0.528 \pm 0.012} & \num{0.523 \pm 0.011} & \num{0.525 \pm 0.016} & \num{0.528 \pm 0.022} & \num{0.513 \pm 0.005} \\
\bottomrule
\end{tabular}
}%
\caption{ROC--AUC on MLP-synthetic datasets at $\{0, 4, 16\}$ shots,
across model and complexity level. Complexity increases from MLP0
(simplest) to MLP9 (most complex).}
\label{tab:synthetic_models_shots}
\end{table*}
% -----------------------------------------------------------------------------

\begin{figure}[ht!]
    \centering
    \includegraphics[width=\linewidth]{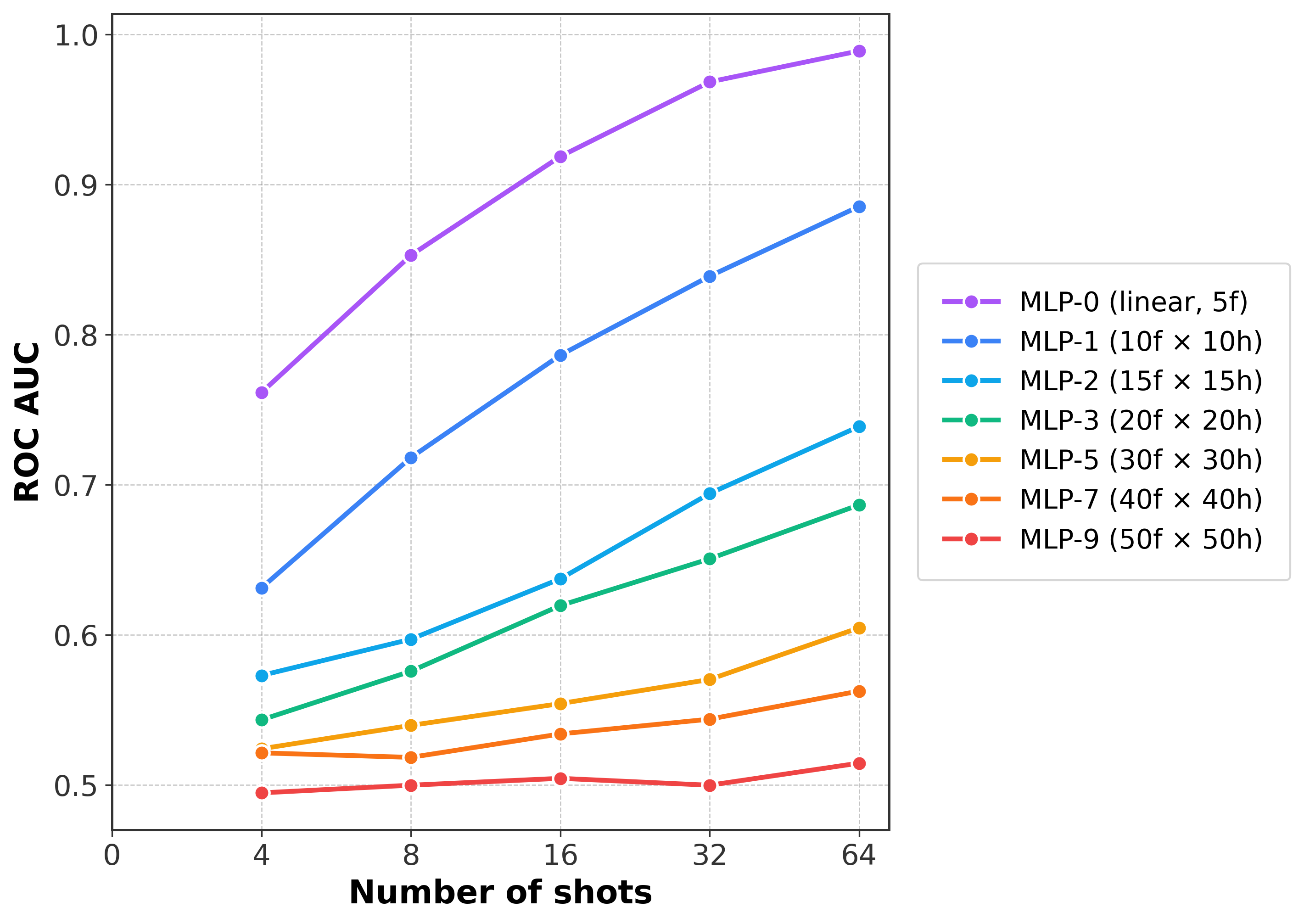}
    \caption{TabPFN-3 Plus ROC--AUC across the seven MLP-prior difficulty
    levels. The curve flattens as complexity rises, defining the upper
    end of the grid we use to test LLMs in
    Section~\ref{sec:q1_h1b} of the main text.}
    \label{fig:rq6_tabpfn_mlp}
    \Description{ROC--AUC on MLP-synthetic datasets at ${0, 4, 16}$ shots, across model and complexity level.}
\end{figure}

\begin{figure}[ht!]
    \centering
    \includegraphics[width=\linewidth]{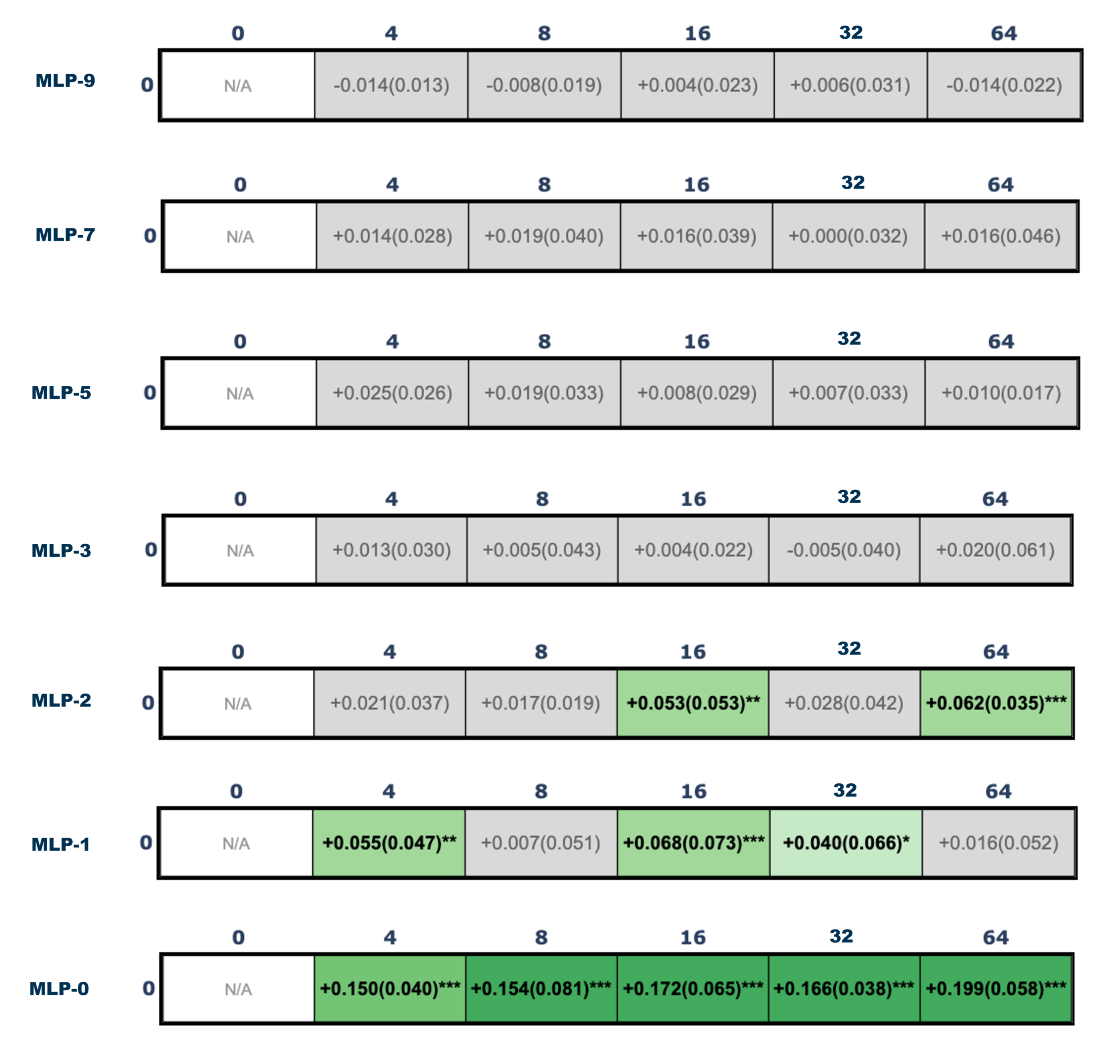}
    \caption{Statistical significance of the per-shot ROC--AUC gain
    over the zero-shot baseline on the MLP-synthetic grid, computed
    via DeLong's test aggregated with Stouffer's method across five
    generation seeds per level. Each cell shows the mean $\Delta$
    (standard deviation in parentheses); green shading marks cells
    significant at $p < 0.05$ ($\ast$), $p < 0.01$ ($\ast\ast$), and
    $p < 0.001$ ($\ast\ast\ast$). Only MLP-0 shows consistent and
    significant gains; from MLP-1 onward, few-shot improvements are
    not statistically distinguishable from zero across the family.}
    \label{fig:rq6_delong_mlp}
    \Description{Statistical significance of the per-shot ROC--AUC gain over the zero-shot baseline on the MLP-synthetic grid, computed via DeLong's test aggregated with Stouffer's method across five generation seeds per level.}
\end{figure}

H1B uses the controlled MLP-synthetic complexity grid
(Table~\ref{tab:datasets_synthetic}) to delineate the level of intrinsic dataset
difficulty at which few-shot demonstrations stop helping.
Figure~\ref{fig:rq6_tabpfn_mlp} shows the TabPFN-3 Plus baseline on this
grid, which serves both as a reference foundation-model and as a
sanity check that the difficulty axis is well-ordered.
Figure~\ref{fig:rq6_delong_mlp} reports the statistical significance
of the per-shot gain over the zero-shot baseline at each MLP level
(DeLong's test combined with Stouffer's method, aggregated across
five generation seeds per level).

% \clearpage

% =============================================================================
% -----------------------------------------------------------------------------

% =============================================================================
\section{Text-Tabular Niche Study (H1c)}
\label{app:texttab}
% =============================================================================

This appendix supports Section~\ref{sec:q1_text}. It gives the full
text-tabular dataset descriptions, per-dataset results for all three
column-level variants, the complete serialization ablation, and the
meta-feature analysis referenced from the main text.

\begin{table*}[!htbp]
\sisetup{
  separate-uncertainty = true,
  table-align-uncertainty = true,
  table-figures-uncertainty = 1,
}
\centering
\footnotesize
\setlength{\tabcolsep}{5pt}
\begin{tabular}{llccccc}
\toprule
Shots & Dataset View & Qwen3-14B & GPT-4o-mini & GLM-4.7 Flash & TabPFN & CatBoost \\
\midrule
\multirow{3}{*}{\textbf{0-shot}} & Full dataset & \num{0.650 \pm 0.108} & \num{0.677 \pm 0.134} & \textemdash & \num{0.500 \pm 0.000} & \num{0.500 \pm 0.000} \\
 & Text only & \num{0.649 \pm 0.105} & \num{0.687 \pm 0.129} & \textemdash & \num{0.500 \pm 0.000} & \num{0.500 \pm 0.000} \\
 & Tabular only & \num{0.629 \pm 0.110} & \num{0.640 \pm 0.111} & \textemdash & \num{0.500 \pm 0.000} & \num{0.500 \pm 0.000} \\
\midrule
\multirow{3}{*}{\textbf{16-shot}} & Full dataset & \num{0.719 \pm 0.126} & \num{0.768 \pm 0.129} & \textemdash & \num{0.719 \pm 0.143} & \num{0.578 \pm 0.010} \\
 & Text only & \num{0.725 \pm 0.122} & \num{0.740 \pm 0.132} & \textemdash & \num{0.539 \pm 0.070} & \num{0.543 \pm 0.027} \\
 & Tabular only & \num{0.699 \pm 0.131} & \num{0.692 \pm 0.121} & \textemdash & \num{0.661 \pm 0.122} & \num{0.605 \pm 0.069} \\
\bottomrule
\end{tabular}
\caption{Mean ROC-AUC $\pm$ standard deviation across the eight text--tabular datasets for full, text-only, and tabular-only views. Since TabPFN and CatBoost do not support label-free inference, their 0-shot entries are fixed to random ROC-AUC $0.5$.}
\label{tab:h1c_text_tabular_views}
\end{table*}

\subsection{Text-tabular datasets}
\label{app:texttab_datasets}

The 8 text-tabular datasets are summarized in Table~\ref{tab:texttab_datasets}.
Each is additionally evaluated as a \emph{text-only} variant (structured
columns removed) and a \emph{tabular-only} variant (text columns removed).
Datasets larger than 2{,}000 rows are subsampled to 1{,}000 rows, preserving
class balance.

\begin{itemize}[noitemsep, leftmargin=*]
    \item \textbf{Real / Fake Job Postings}: binary classification of job
    listings as fraudulent vs.\ genuine; text columns include the title,
    description, and requirements.
    \item \textbf{California Real Estate Market 2026}: predict the property
    type, binarized from the multi-class label; one free-form description
    column alongside structured listing features. Released in 2026 and
    therefore unavailable to LLM pretraining as of the models' cutoffs.
    \item \textbf{Wine Tasting Reviews}: binarized wine-quality score from
    review text and structured wine metadata.
    \item \textbf{Amazon Fine Food Reviews}: binarized review score (low vs.\
    high); text columns include the review body and summary.
    \item \textbf{Fake Reviews Dataset}: classifier of computer-generated
    vs.\ genuine product reviews; a single review-text column with one
    numerical and one categorical feature.
    \item \textbf{PubMed MeSH}: a single MeSH term reduced to a binary topic
    indicator; abstracts and titles are the text columns.
    \item \textbf{Food Recipes}: binarized recipe category from ingredients,
    instructions, and recipe metadata.
    \item \textbf{Online Teaching Feedback}: sentiment-derived binary target
    from student feedback text and lecture metadata.
\end{itemize}

\begin{table*}[t]
\centering
\small
\caption{Text-tabular dataset collection. Datasets larger than 4{,}000 rows
were subsampled to 1{,}000 stratified by label (marked ``curated to 1,000'');
smaller datasets were kept at their original size. Column-type counts are
reported per the source.}
\label{tab:texttab_datasets}
\begin{tabular}{lcccccc}
\toprule
Dataset & Source & \#num & \#cat & \#text & Working size & Original \\
\midrule
fake\_job\_postings & Kaggle & 1 & 7 & 8 & curated to 1,000 & 18,000 \\
california\_price & Kaggle & 9 & 2 & 1 & curated to 1,000 & 9,500 \\
wine\_testing\_reviews & Kaggle & 2 & 4 & 8 & curated to 1,000 & 130,000 \\
reviews\_amazon & Kaggle & 3 & 1 & 2 & curated to 1,000 & 570,000 \\
pubmed & Kaggle & 0 & 4 & 2 & curated to 1,000 & 50,000 \\
recipes & Kaggle & 14 & 5 & 9 & 1,000 (original) & 1,000 \\
fake\_reviews & Kaggle & 1 & 1 & 1 & curated to 1,000 & 40,000 \\
online\_teaching & Kaggle & 5 & 0 & 1 & 1,400 (original) & 1,400 \\
\bottomrule
\end{tabular}
\end{table*}

\begin{figure}[t]
  \centering
  \includegraphics[width=\linewidth]{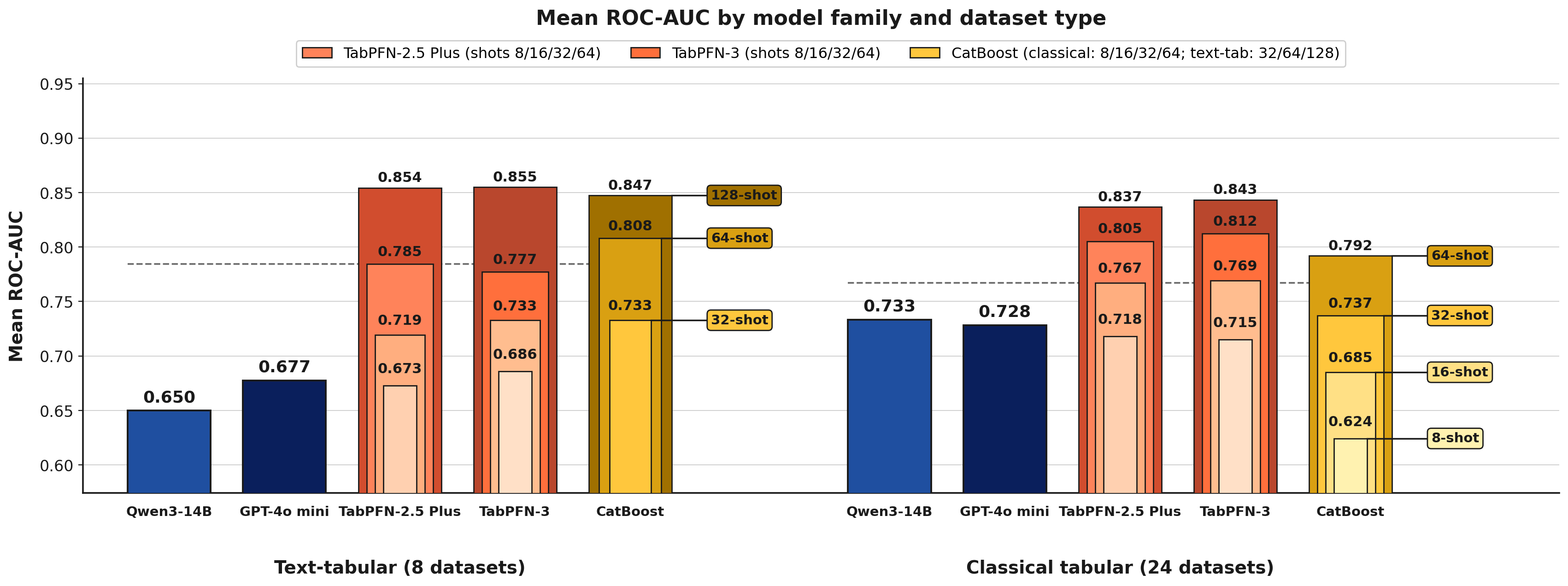}
  \caption{Mean zero-shot ROC-AUC on the 8 text-tabular and 24 classical
  datasets, LLMs vs.\ TabPFN-2.5/3 Plus and CatBoost at increasing shot
  budgets. Dashed reference line marks the label budget at which each
  baseline family first matches the zero-shot LLMs. Referenced from
  Section~\ref{sec:q1_text}.}
  \label{fig:texttab_headline}
\end{figure}

\subsection{Per-dataset results}
\label{app:texttab_results}

Tables~\ref{tab:texttab_full}, \ref{tab:texttab_text}, and
\ref{tab:texttab_tabular} report per-dataset zero-shot ROC-AUC for the
\emph{full}, \emph{text-only}, and \emph{tabular-only} variants respectively,
underlying the aggregates in Section~\ref{sec:q1_text}.

\begin{table*}[t]
\centering
\small
\caption{Per-dataset ROC-AUC on the 8 text-tabular datasets, \emph{full}
variant. LLMs zero-shot; TabPFN-2.5/3 Plus at 16 shots; CatBoost at 32 shots
(CatBoost was not run at lower shots for text-tabular datasets).}
\label{tab:texttab_full}
\resizebox{\linewidth}{!}{%
\begin{tabular}{lccccccc}
\toprule
Dataset & Qwen3-1.7B & Qwen3-8B & Qwen3-14B & GPT-4o-mini & TabPFN-2.5 (16) & TabPFN-3 (16) & CatBoost (32) \\
\midrule
fake\_job\_postings & 0.536 & 0.536 & 0.541 & 0.623 & 0.888 & 0.895 & --- \\
california\_price & 0.527 & 0.790 & 0.791 & 0.796 & 0.661 & 0.715 & 0.698 \\
wine\_testing\_reviews & 0.724 & 0.788 & 0.773 & 0.820 & 0.685 & 0.703 & 0.688 \\
reviews\_amazon & 0.618 & 0.690 & 0.698 & 0.685 & 0.674 & 0.701 & --- \\
fake\_reviews & 0.541 & 0.643 & 0.619 & 0.565 & 0.505 & 0.509 & 0.604 \\
pubmed & 0.544 & 0.526 & 0.743 & 0.886 & 0.674 & 0.677 & 0.713 \\
recipes & 0.581 & 0.603 & 0.517 & 0.507 & 0.667 & 0.666 & 0.727 \\
online\_teaching & 0.515 & 0.525 & 0.517 & 0.536 & 1.000 & 0.997 & 0.966 \\
\midrule
Mean & 0.573 & 0.638 & 0.650 & 0.677 & 0.719 & 0.733 & 0.733 \\
\bottomrule
\end{tabular}
}%
\end{table*}

\begin{table*}[t]
\centering
\small
\caption{Per-dataset ROC-AUC on the 8 text-tabular datasets, \emph{text-only}
variant. LLMs zero-shot; TabPFN-2.5/3 Plus at 16 shots; CatBoost at 32 shots.}
\label{tab:texttab_text}
\resizebox{\linewidth}{!}{%
\begin{tabular}{lccccccc}
\toprule
Dataset & Qwen3-1.7B & Qwen3-8B & Qwen3-14B & GPT-4o-mini & TabPFN-2.5 (16) & TabPFN-3 (16) & CatBoost (32) \\
\midrule
fake\_job\_postings & 0.568 & 0.583 & 0.520 & 0.538 & 0.623 & 0.667 & --- \\
california\_price & 0.563 & 0.695 & 0.736 & 0.786 & 0.503 & 0.502 & 0.507 \\
wine\_testing\_reviews & 0.562 & 0.672 & 0.770 & 0.805 & 0.500 & 0.500 & 0.553 \\
reviews\_amazon & 0.607 & 0.690 & 0.706 & 0.699 & 0.501 & 0.502 & --- \\
fake\_reviews & 0.535 & 0.669 & 0.625 & 0.609 & 0.500 & 0.500 & 0.610 \\
pubmed & 0.529 & 0.551 & 0.778 & 0.904 & 0.493 & 0.507 & 0.609 \\
recipes & 0.535 & 0.574 & 0.555 & 0.642 & 0.691 & 0.664 & 0.515 \\
online\_teaching & 0.508 & 0.503 & 0.505 & 0.510 & 0.498 & 0.497 & 0.501 \\
\midrule
Mean & 0.551 & 0.617 & 0.649 & 0.687 & 0.539 & 0.542 & 0.549 \\
\bottomrule
\end{tabular}
}%
\end{table*}

\begin{table*}[t]
\centering
\small
\caption{Per-dataset ROC-AUC on the 8 text-tabular datasets,
\emph{tabular-only} variant. LLMs zero-shot; TabPFN-2.5/3 Plus at 16 shots;
CatBoost at 32 shots.}
\label{tab:texttab_tabular}
\resizebox{\linewidth}{!}{%
\begin{tabular}{lccccccc}
\toprule
Dataset & Qwen3-1.7B & Qwen3-8B & Qwen3-14B & GPT-4o-mini & TabPFN-2.5 (16) & TabPFN-3 (16) & CatBoost (32) \\
\midrule
fake\_job\_postings & 0.538 & 0.577 & 0.568 & 0.669 & 0.919 & 0.897 & 0.841 \\
california\_price & 0.646 & 0.799 & 0.750 & 0.755 & 0.646 & 0.715 & 0.729 \\
wine\_testing\_reviews & 0.752 & 0.799 & 0.821 & 0.849 & 0.722 & 0.709 & 0.715 \\
reviews\_amazon & 0.500 & 0.727 & 0.726 & 0.682 & 0.682 & 0.677 & 0.693 \\
fake\_reviews & 0.507 & 0.508 & 0.526 & 0.522 & 0.505 & 0.503 & 0.504 \\
pubmed & --- & --- & 0.520 & 0.556 & 0.667 & 0.673 & 0.709 \\
recipes & 0.528 & 0.523 & 0.562 & 0.549 & 0.640 & 0.640 & 0.675 \\
online\_teaching & 0.526 & 0.545 & 0.555 & 0.537 & 0.508 & 0.506 & 0.490 \\
\midrule
Mean & 0.571 & 0.640 & 0.629 & 0.640 & 0.661 & 0.665 & 0.670 \\
\bottomrule
\end{tabular}
}%
\end{table*}

\subsection{Serialization ablation}
\label{app:texttab_serialization}

Table~\ref{tab:serialization_texttab} reports the mean serialization rank
(1 = best, 5 = worst) on the combined pool of 32 datasets (24 classical + 8
text-tabular \emph{full}), supporting the H2d discussion in
Section~\ref{sec:q2_h2d}. Tables~\ref{tab:serialization_classical}--\ref{tab:serialization_tabularonly}
break the same ranking down by pool and column-level variant.

\begin{table}[t]
\centering
\small
\caption{Mean serialization rank on the combined pool of 32 datasets (24
classical + 8 text-tabular \emph{full}), zero-shot. Lower mean is better;
``First (\%)'' is the share of datasets on which the format wins.}
\label{tab:serialization_texttab}
\begin{tabular}{lcccc}
\toprule
& \multicolumn{2}{c}{GPT-4o-mini} & \multicolumn{2}{c}{Qwen3-14B} \\
Serialization & Mean & First (\%) & Mean & First (\%) \\
\midrule
markdown & \textbf{2.19} & 25.0 & 2.91 & 15.6 \\
html & 2.62 & 21.9 & \textbf{2.38} & 21.9 \\
feat\_val & 2.75 & \textbf{34.4} & 2.56 & \textbf{31.2} \\
markdown\_masked & 3.81 & 6.2 & 3.31 & 18.8 \\
feat\_val\_masked & 3.62 & 12.5 & 3.81 & 12.5 \\
\bottomrule
\end{tabular}
\end{table}

\begin{table}[t]
\centering
\small
\caption{Mean serialization rank, 24 classical benchmarks only.}
\label{tab:serialization_classical}
\begin{tabular}{lcccc}
\toprule
& \multicolumn{2}{c}{GPT-4o-mini} & \multicolumn{2}{c}{Qwen3-14B} \\
Format & Mean & First (\%) & Mean & First (\%) \\
\midrule
feat\_val & 2.29 & 45.8 & 2.25 & 37.5 \\
feat\_val\_masked & 3.79 & 8.3 & 4.04 & 8.3 \\
html & 2.62 & 20.8 & 2.25 & 25.0 \\
markdown & 2.08 & 20.8 & 2.96 & 12.5 \\
markdown\_masked & 4.21 & 4.2 & 3.46 & 16.7 \\
\bottomrule
\end{tabular}
\end{table}

\begin{table}[t]
\centering
\small
\caption{Mean serialization rank, 8 text-tabular \emph{text-only} variant.}
\label{tab:serialization_textonly}
\begin{tabular}{lcccc}
\toprule
& \multicolumn{2}{c}{GPT-4o-mini} & \multicolumn{2}{c}{Qwen3-14B} \\
Format & Mean & First (\%) & Mean & First (\%) \\
\midrule
feat\_val & 2.75 & 12.5 & 2.88 & 37.5 \\
feat\_val\_masked & 3.88 & 12.5 & 3.62 & 0.0 \\
html & \textbf{2.38} & \textbf{37.5} & \textbf{2.38} & \textbf{37.5} \\
markdown & 3.50 & 12.5 & 2.25 & 25.0 \\
markdown\_masked & 2.50 & 25.0 & 3.88 & 0.0 \\
\bottomrule
\end{tabular}
\end{table}

\begin{table}[t]
\centering
\small
\caption{Mean serialization rank, 8 text-tabular \emph{tabular-only} variant.}
\label{tab:serialization_tabularonly}
\begin{tabular}{lcccc}
\toprule
& \multicolumn{2}{c}{GPT-4o-mini} & \multicolumn{2}{c}{Qwen3-14B} \\
Format & Mean & First (\%) & Mean & First (\%) \\
\midrule
feat\_val & 3.00 & \textbf{37.5} & 2.50 & \textbf{37.5} \\
feat\_val\_masked & 3.38 & 12.5 & 3.38 & 12.5 \\
html & 3.25 & 12.5 & \textbf{2.25} & \textbf{37.5} \\
markdown & \textbf{2.38} & 12.5 & 3.62 & 0.0 \\
markdown\_masked & 3.00 & 25.0 & 3.25 & 12.5 \\
\bottomrule
\end{tabular}
\end{table}

\subsection{Memorization check on the text-tabular suite}
\label{app:texttab_memorization}

As with the core suite (Appendix~\ref{app:rq1}), we cannot fully rule out
that LLMs retain some memorized signal about the underlying datasets from
pretraining~\cite{bordt2024elephants}. We rely on two mitigations: the
\texttt{california\_price} dataset was released in 2026 and is therefore
unavailable to any evaluated model's pretraining corpus, and it shows no
systematic advantage over the other 7 text-tabular datasets in the
main-text comparisons. Combined with the recognition/reconstruction probes
already reported for the core suite in Appendix~\ref{app:rq1}, and with the
prior zero-shot-vs-post-cutoff comparison on the core suite
(Section~\ref{sec:q1_h1a}), this suggests the text-tabular niche identified
in Section~\ref{sec:q1_text} reflects a structural property of the
column-type mix rather than dataset-specific memorization, though residual
per-dataset contamination cannot be excluded given the suite's size ($n=8$).

\end{document}